\documentclass[11pt]{article}

\PassOptionsToPackage{table}{xcolor}
\usepackage[preprint]{acl}

\usepackage[utf8]{inputenc}
\usepackage[T1]{fontenc}
\usepackage{amsmath,amssymb,amsfonts}
\usepackage{graphicx}
\usepackage{booktabs}
\usepackage{multirow}
\usepackage{subcaption}
\usepackage{enumitem}
\usepackage{tabularx}
\usepackage{adjustbox}
\usepackage{placeins}
\usepackage{times}
\usepackage{latexsym}

\newcommand{\method}{\textsc{Compress-Distill}}
\newcommand{\think}{\texttt{<think>}}
\newcommand{\thinkend}{\texttt{</think>}}

\title{Compress-Distill: Reasoning Trace Compression for\\Efficient Knowledge Distillation}

\author{
  Maxime Griot\textsuperscript{1,$\dagger$}, 
  Paul Steven Scotti\textsuperscript{2},
  Tanishq Mathew Abraham\textsuperscript{2}
  \\
  \textsuperscript{1}Université catholique de Louvain,
  \textsuperscript{2}Sophont Inc.
}

\begin{document}
\maketitle

\begingroup
\renewcommand{\thefootnote}{$\dagger$}
\footnotetext{This work was performed while serving as an independent consultant for Sophont Inc.}
\endgroup

\begin{abstract}
Reasoning models produce long chain-of-thought traces that are costly to distill and encourage verbose student outputs. We study post-hoc compression of such traces before knowledge distillation. Two teachers, Qwen3.5-397B-A17B and gpt-oss-120B, generate about 283k correct traces each; two instruction-tuned models then compress them to 8.6--21.0\% of their original character length. Across a 48-run main grid plus seven Qwen-teacher truncation ablations, compressed traces reduce training tokens to 12--30\% of raw, speed up training by 2.0--7.6$\times$, and shorten inference outputs by 3--19$\times$ with smaller reductions under the shorter gpt-oss teacher. However, raw traces retain the highest downstream accuracy at every scale and for both teachers. A length-matched raw-trace truncation ablation shows that compression is not merely benefiting from a smaller token budget: model-compressed traces usually beat or match naive truncation, especially for smaller students, while maintaining shorter inference outputs. Overall, reasoning-trace compression offers an accuracy--efficiency trade-off rather than a free improvement: students retain up to 96\% of raw-trace accuracy while gaining up to 18$\times$ higher per-token efficiency, and at the 0.8B scale under LoRA compressed traces narrow the raw-vs-compressed gap but do not exceed raw.
\end{abstract}

% ===================================================================
\section{Introduction}
\label{sec:intro}
% ===================================================================

Reasoning-tuned models such as Qwen3 \citep{qwen2025qwen3} and DeepSeek-R1 \citep{deepseek2025r1} often produce long chain-of-thought (CoT) traces. These traces can improve step-by-step reasoning \citep{wei2022chain,kojima2022large}, but they also increase supervised fine-tuning cost and teach students to generate verbose, expensive outputs.

We ask whether teacher reasoning can be compressed before distillation, and what accuracy--efficiency trade-off this creates. Our pipeline generates correct traces with a large teacher, rewrites them with instruction-tuned compressor models, and fine-tunes students on raw, compressed, or answer-only targets. Across two teachers, four students, two training regimes, a 48-run main grid, and seven additional Qwen-teacher truncation ablations, we evaluate downstream accuracy, training cost, inference length, truncation, and per-token efficiency.

We find that compression sharply reduces cost but underperforms raw traces at every evaluated scale and under both teachers. Still, compression provides a worthwhile efficiency trade-off: compressed traces cut training tokens to 12--30\% of raw, reduce wall-clock training by 2.0--7.6$\times$, and shorten inference reasoning by 3--19$\times$. The answer-only baseline is cheapest but performs worst and is unstable under full fine-tuning.

Our contribution is an empirical characterization of reasoning-trace compression as a practical trade-off rather than a free improvement. We provide:
\begin{enumerate}[leftmargin=*,topsep=2pt,itemsep=1pt]
  \item A pipeline for trace generation, compression, and student fine-tuning.
  \item A compression analysis across teachers, domains, and compressor models.
  \item A 48-run study reporting training cost, downstream accuracy, inference length, truncation, and per-token efficiency.
  \item A length-matched truncation ablation testing whether compressed traces help beyond a reduced training-token budget.
  \item An analysis of failure modes, including small-student termination failures and unstable answer-only training.
\end{enumerate}

% ===================================================================
\section{Related Work}
\label{sec:related}
% ===================================================================

\paragraph{Chain-of-thought reasoning.}
\citet{wei2022chain} showed that prompting LLMs to produce step-by-step reasoning improves arithmetic, commonsense, and symbolic reasoning.
\citet{kojima2022large} showed zero-shot CoT prompting elicits reasoning without task-specific examples.
Reasoning-tuned models such as DeepSeek-R1 \citep{deepseek2025r1} and QwQ \citep{qwen2025qwq} are explicitly trained to emit long deliberation traces.

\paragraph{Knowledge distillation for reasoning.}
\citet{ho2023large}, \citet{magister2023teaching}, and \citet{shridhar2023distilling} all show that fine-tuning smaller models on traces produced by a larger teacher transfers reasoning ability, typically using the teacher's full verbose trace as the target.
We instead study a step in between: rewriting the teacher's trace with a separate compressor before distillation.

\paragraph{Reasoning trace optimisation.}
Several works seek to reduce the cost of long reasoning traces. \citet{xia2025tokenskip} prune low-importance CoT tokens and fine-tune models to learn controllable shortcuts, while \citet{kang2025c3ot} use a stronger model to generate shorter CoTs for training. Other approaches control reasoning length through token-budget estimation, RL, pruning, or self-training \citep{han2024token,aggarwal2025l,luo2025o1,munkhbat2025self}, and \citet{zhang2026tokensqueeze} explicitly studies performance-preserving reasoning compression. Our work is closest in spirit to this line, but differs in evaluating post-hoc model-based rewriting of verified teacher traces as a distillation data intervention across two teachers, four students, and both LoRA and full fine-tuning.

\paragraph{Inference-time efficiency for reasoning.}
A complementary line targets the decoding budget directly, with overthinking analyses, adaptive slow/fast reasoning, and sketch-like rationales \citep{chen2025do,shen2025dast,aytes2025sketch,xu2025chain,yang2025think}. We keep greedy decoding and an 8{,}192-token cap fixed to isolate the training-time intervention. The regimes are complementary: training-time compression pays a one-time data cost for shorter default outputs, while decoding-time controls are reusable across checkpoints but always-on at inference.

% ===================================================================
\section{Method}
\label{sec:method}
% ===================================================================

\method{} is a three-stage pipeline; each stage is independently resumable and writes JSONL outputs that feed the next stage.

% -------------------------------------------------------------------
\subsection{Stage 1: Trace Generation}
\label{sec:trace_gen}
% -------------------------------------------------------------------

Given a teacher $\mathcal{M}_T$ and benchmark questions $\{q_i\}_{i=1}^N$, we sample a response $r_i = \mathcal{M}_T(q_i)$ that contains a reasoning trace $t_i$ (within \think{}\ldots\thinkend{}) and a final answer $a_i$.
We extract $a_i$ using a type-specific cascade and verify against the canonical answer $a_i^*$.
If $a_i \neq a_i^*$, we resample up to $K$ times.
Only correct traces are retained.

Answers are extracted with a type-specific cascade (\verb|\boxed{}|, \verb|####|, ``the answer is'', or fallbacks for numbers/letters/booleans/\LaTeX{}) and matched against $a_i^*$.
All pending questions are submitted in a single batch to vLLM~\citep{kwon_efficient_2023}, and questions answered correctly are removed before the next rejection-sampling round.

% -------------------------------------------------------------------
\subsection{Stage 2: Trace Compression}
\label{sec:compression}
% -------------------------------------------------------------------

For each correct triple $(q_i, t_i, a_i)$ from Stage~1, we apply a compressor $\mathcal{M}_C$ with a single generic prompt that asks for a shorter version of the reasoning preserving the essential logical steps, key insights, and the final answer. The compressed trace $\hat{t}_i = \mathcal{M}_C(q_i, t_i, a_i)$ is generated with low temperature ($\tau{=}0.3$). We measure the per-example character-level compression ratio $\rho_i = |\hat{t}_i| / |t_i|$. We run two compressors independently, producing two parallel compressed datasets that differ in compression aggressiveness and style.

% -------------------------------------------------------------------
\subsection{Stage 3: Student Training}
\label{sec:training}
% -------------------------------------------------------------------

We fine-tune student models $\mathcal{M}_S$ on raw traces, compressed traces, or an \emph{answer-only} ablation that removes the entire \think{}\ldots\thinkend{} block, using a chat template:
\[
  \text{User: } q_i \quad\rightarrow\quad
  \text{Asst: } \langle\text{think}\rangle\, \tilde{t}_i \,\langle/\text{think}\rangle\; a_i,
\]
where $\tilde{t}_i \in \{t_i, \hat{t}_i, \varnothing\}$ (the last replaces the think block with the empty string, leaving only the final answer).
Training uses next-token prediction on assistant tokens only, with sample packing.

We evaluate a grid over two teachers $\mathcal{M}_T$ (Qwen3.5-397B-A17B and gpt-oss-120B), four students (Qwen3.5-0.8B-Base, Qwen3.5-9B-Base, Llama-3.1-8B, and gpt-oss-20B), four data sources (raw, Llama-70B-compressed, Ministral-14B-compressed, and answer-only), and two one-epoch methods (LoRA and full / FSDP fine-tuning)~\citep{hu2022lora}. Answer-only is run only once because it uses no teacher trace and is omitted for the gpt-oss-20B student for the same reason. This gives 48 main-grid runs: 27 under the Qwen teacher and 21 under the gpt-oss teacher.

To test whether compressed traces help because they are better supervision rather than simply shorter supervision, we add a \emph{traces-truncated} ablation for the Qwen-teacher runs. For each example, we tokenize the raw Qwen trace with the Ministral tokenizer and truncate it to the token length of the corresponding Ministral-compressed trace, allowing a $\pm1$ token mismatch after decode/re-encode. The final answer is kept unchanged. This produces a compute-equivalent raw-trace baseline at the same per-example training-token budget as the most aggressive compressor. We train the same seven Qwen-teacher student/method configurations as the reasoning-trace grid.

% ===================================================================
\section{Experimental Setup}
\label{sec:experiments}
% ===================================================================

\begin{table*}[t]
  \centering
  \begin{adjustbox}{max width=\textwidth}
\begin{tabular}{lllllrr}
\toprule
\textbf{Domain} & \textbf{Dataset} & \textbf{Answer Type} & \textbf{Train Split} & \textbf{Eval Split} & \textbf{$|$Train$|$} & \textbf{$|$Eval$|$} \\
\midrule
\multirow{4}{*}{Mathematics}
  & GSM8k~\citep{cobbe2021trainingverifierssolvemath}                    & Numeric & train & test                & 7{,}473   & 1{,}319 \\
  & MultiArith~\citep{roy-roth-2015-solving}               & Numeric & train & test                & 420       & 180 \\
  & SVAMP~\citep{patel-etal-2021-nlp}                    & Numeric & train & ---                 & 1{,}000   & --- \\
  & AQUA-RAT~\citep{ling2017programinductionrationalegeneration}                 & Letter  & train & test                & 97{,}467  & 254 \\
\midrule
\multirow{2}{*}{Science}
  & ARC-Challenge~\citep{clark2018thinksolvedquestionanswering}            & Letter  & train & test                & 1{,}119   & 1{,}172 \\
  & GPQA Diamond~\citep{rein2024gpqa}             & Letter  & train & ---                 & 198       & --- \\
\midrule
Logic/Commonsense
  & CommonsenseQA~\citep{talmor-etal-2019-commonsenseqa}            & Letter  & train & validation          & 9{,}741   & 1{,}221 \\
\midrule
\multirow{3}{*}{Medical}
  & MedQA~\citep{jin_what_2021}                    & Letter  & train & test                & 10{,}178  & 1{,}273 \\
  & MedMCQA~\citep{pal_medmcqa_2022}                  & Letter  & train & validation          & 182{,}822 & 4{,}183 \\
  & MedExpQA~\citep{Alonso_2024}                 & Letter  & train & test                & 434       & 125 \\
\midrule
MMLU-STEM    & 9 subjects~\citep{hendrycks2021measuring}    & Letter  & dev   & test                & 45        & 1{,}274 \\
MMLU-Medical & 7 subjects~\citep{hendrycks2021measuring}    & Letter  & dev   & test                & 35        & 1{,}417 \\
\bottomrule
\end{tabular}
\end{adjustbox}

  \caption{In-distribution benchmark datasets used in \method{}.}
  \label{tab:benchmarks}
\end{table*}

Trace generation and downstream evaluation use the in-distribution datasets in Table~\ref{tab:benchmarks}: ten datasets across four domains plus two MMLU subject groups~\citep{hendrycks2021measuring}. SVAMP~\citep{patel-etal-2021-nlp} and GPQA Diamond~\citep{rein2024gpqa} lack labeled held-out splits and are used for trace generation only. Downstream evaluation also includes OOD-reasoning datasets (HellaSwag~\citep{zellers_hellaswag_2019}, WinoGrande~\citep{sakaguchi_winogrande_2021}, PIQA~\citep{bisk_piqa_2020}, ARC-Easy~\citep{clark2018thinksolvedquestionanswering}, BoolQ~\citep{clark_boolq_2019}, TruthfulQA~\cite{lin-etal-2022-truthfulqa}) and OOD-knowledge groups (MMLU humanities, social sciences, other).

All training uses Axolotl with BF16, FlashAttention~2~\citep{dao2024flashattention}, CutCrossEntropy~\cite{wijmans2025cut}, 16{,}384-token sequences with sample packing, and a 5\% validation split. LoRA runs use rank 64, $\alpha{=}32$, dropout 0.05, all attention and MLP projections, LR $1\!\times\!10^{-4}$ with cosine schedule and 10\% warmup, one epoch, and 8-bit AdamW. Full fine-tuning uses LR $2\!\times\!10^{-5}$ for one epoch; the 8B, 9B, and 20B students use FSDP v2 with full sharding and activation checkpointing, while the 0.8B model fits on a single GPU.

Each student is evaluated with the same chat template used during training. Generation is greedy with an 8{,}192-token cap. We report per-group accuracy for Math, Science, Medicine, Common-sense, OOD-Reason., and OOD-Knownledge. Overall accuracy across all records, inference output length, and per-token efficiency metric are defined in Section~\ref{sec:results_eval}.

% ===================================================================
\section{Results}
\label{sec:results}
% ===================================================================

% -------------------------------------------------------------------
\subsection{Trace Generation}
\label{sec:results_trace}
% -------------------------------------------------------------------

The Qwen3.5-397B-A17B teacher yields 283{,}335 verified-correct traces across the trace-generation pool. Accuracy and verbosity vary widely: on easy arithmetic (MultiArith, 99.0\%) the teacher emits short traces (mean 1{,}750 characters), while on graduate-level science (GPQA Diamond, 68.2\%) it averages 18{,}734-character traces---more than 10$\times$ longer. This per-problem variance is what motivates compression.

The gpt-oss-120B teacher (high reasoning) answers 281{,}911 questions correctly with traces that are roughly half as long character-for-character (e.g., MultiArith mean 716, GPQA Diamond mean 13{,}407): similar accuracy as Qwen3.5-397B-A17B, substantially less verbosity. The two teachers thus probe opposite ends of the verbosity axis, which determines how much room downstream compression has to work. Per-dataset statistics for both teachers are in Appendix~\ref{app:trace_generation}, Table~\ref{tab:trace_generation}.

% -------------------------------------------------------------------
\subsection{Compression}
\label{sec:results_compression}
% -------------------------------------------------------------------

\begin{figure*}[t]
  \centering
  \includegraphics[width=\textwidth]{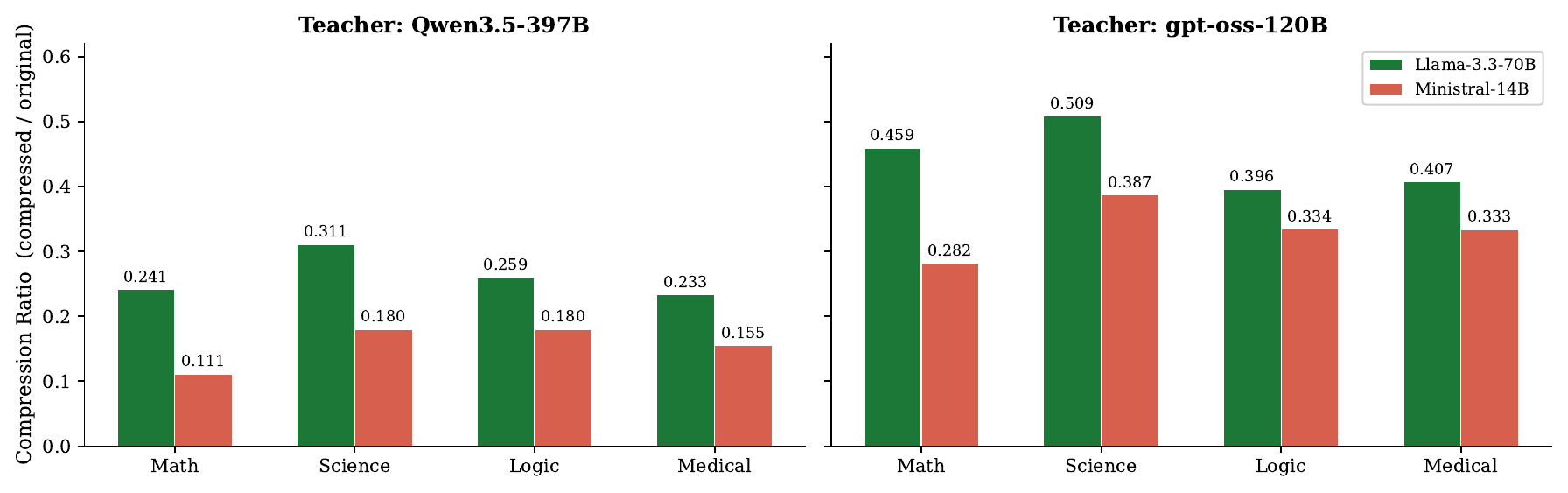}
  \caption{Mean compression ratio by domain (lower is more aggressive), one panel per teacher. The gpt-oss teacher's compressed traces compress to a higher ratio (less compression) than the Qwen teacher's at every (compressor, domain) combination because the source traces are already substantially shorter.}
  \label{fig:compression_ratio}
\end{figure*}

Table~\ref{tab:compression} and Figure~\ref{fig:compression_ratio} summarize compression for each (teacher, compressor) pair; per-dataset statistics are reported in Appendix~\ref{app:compression}, Table~\ref{tab:compression_full}. Under the Qwen teacher, Llama-3.3-70B-Instruct~\citep{grattafiori2024llama3} compresses traces to a mean ratio of $\rho{=}0.142$, while Ministral-3-14B-Instruct-2512~\citep{liu2026ministral3} is more aggressive at $\rho{=}0.086$. Under the gpt-oss teacher~\citep{openai2025gptoss120bgptoss20bmodel}, compression is milder, with mean ratios of $\rho{=}0.210$ for Llama-70B and $\rho{=}0.147$ for Ministral-14B. Thus, Ministral-14B is consistently the more aggressive compressor, while gpt-oss traces retain higher compression ratios because the raw teacher traces are already substantially shorter.

\begin{figure*}[t]
  \centering
  \includegraphics[width=\textwidth]{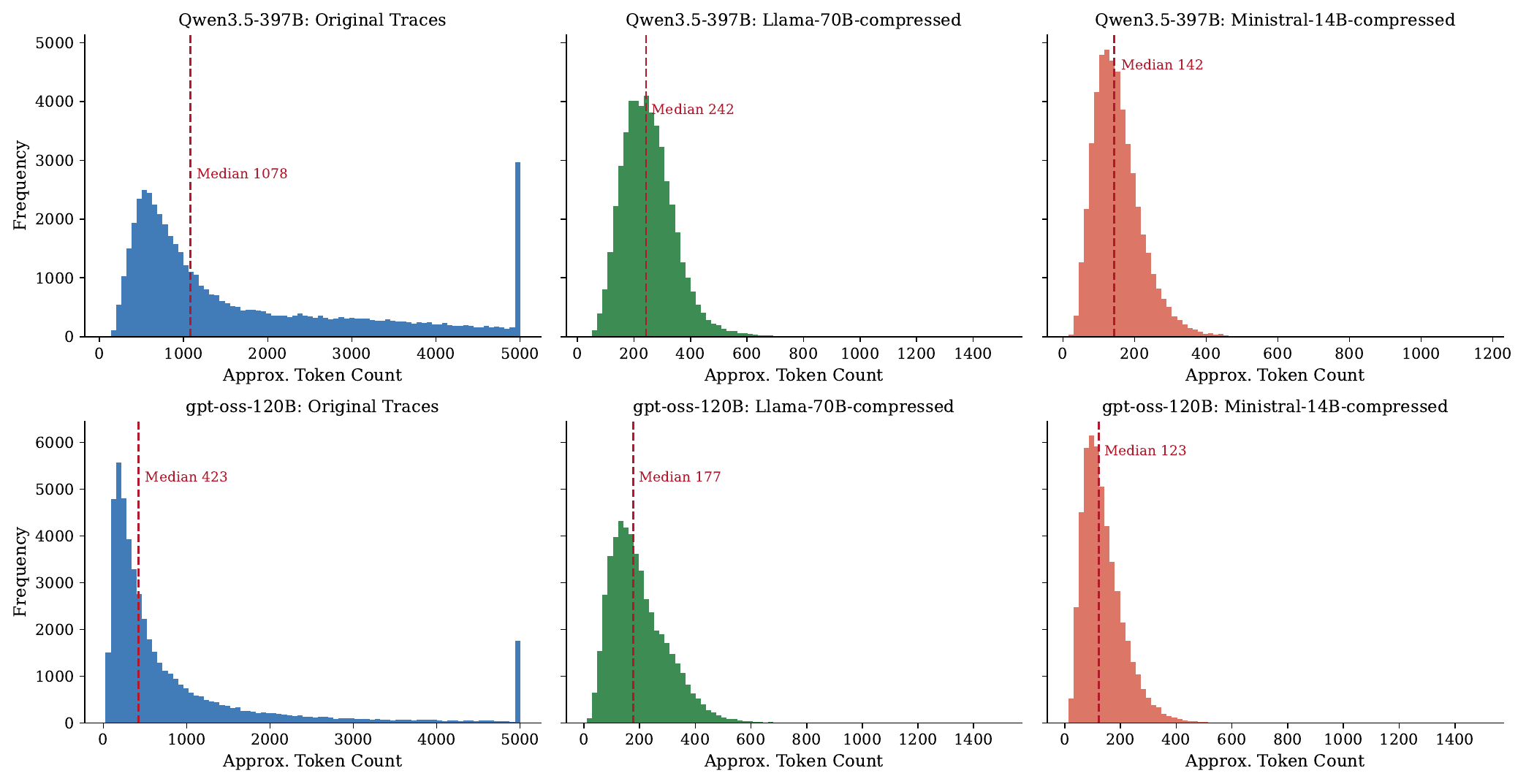}
  \caption{Approximate reasoning token counts: original (left), Llama-70B-compressed (centre), Ministral-14B-compressed (right). Top row: Qwen3.5-397B teacher. Bottom row: gpt-oss-120B teacher. The gpt-oss teacher's original-trace distribution is already concentrated at low token counts; compression sharpens the peak but moves it less in absolute terms than under the Qwen teacher.}
  \label{fig:token_distribution}
\end{figure*}

Compression is strongly dataset-dependent. GPQA Diamond compresses hardest, especially under the Qwen teacher ($\rho{=}0.089$ for Llama-70B and $\rho{=}0.054$ for Ministral-14B), whereas shorter/easier datasets such as MultiArith and ARC leave less room for compression. This pattern suggests that longer reasoning traces contain more removable deliberation. Figure~\ref{fig:token_distribution} shows that compression removes much of the long right tail: after compression, all but the longest 10\% of traces are shorter than 500 estimated tokens under both compressors, although rare long outliers remain.

Under the gpt-oss teacher, the qualitative picture is broadly similar: GPQA Diamond compresses hardest, Ministral-14B is more aggressive than Llama-70B across datasets, and the shortest/easiest datasets leave the least room for compression. However, the absolute amount removed is smaller because there is less redundancy to remove from a teacher whose raw traces are already $\sim$2$\times$ shorter. Consequently, the compressors do not bring the two teachers to a common target length: gpt-oss compressed traces remain shorter in absolute terms, but their compression ratios are higher than under the more verbose Qwen teacher.

\begin{table}[t]
  \centering
  \small
  \begin{tabular}{llrr}
\toprule
\textbf{Teacher} & \textbf{Compressor} & \textbf{Length} & \textbf{$\rho$} \\
\midrule
Qwen3.5-397B & Llama-3.3-70B & 1{,}008 & 0.142 \\
Qwen3.5-397B & Ministral-3-14B & 613 & 0.086 \\
gpt-oss-120B & Llama-3.3-70B & 793 & 0.210 \\
gpt-oss-120B & Ministral-3-14B & 554 & 0.147 \\
\bottomrule
\end{tabular}

  \caption{Aggregate compression statistics. Length is mean compressed length; $\rho$ is mean compressed / original length.}
  \label{tab:compression}
\end{table}

% -------------------------------------------------------------------
\subsection{Training Loss and Efficiency}
\label{sec:results_training}
% -------------------------------------------------------------------

Figure~\ref{fig:loss_curves} shows the loss curves; full wall-clock and token counts are reported in Appendix~\ref{app:training_efficiency}, Table~\ref{tab:efficiency}.

\begin{figure*}[t]
  \centering
  \includegraphics[width=\textwidth]{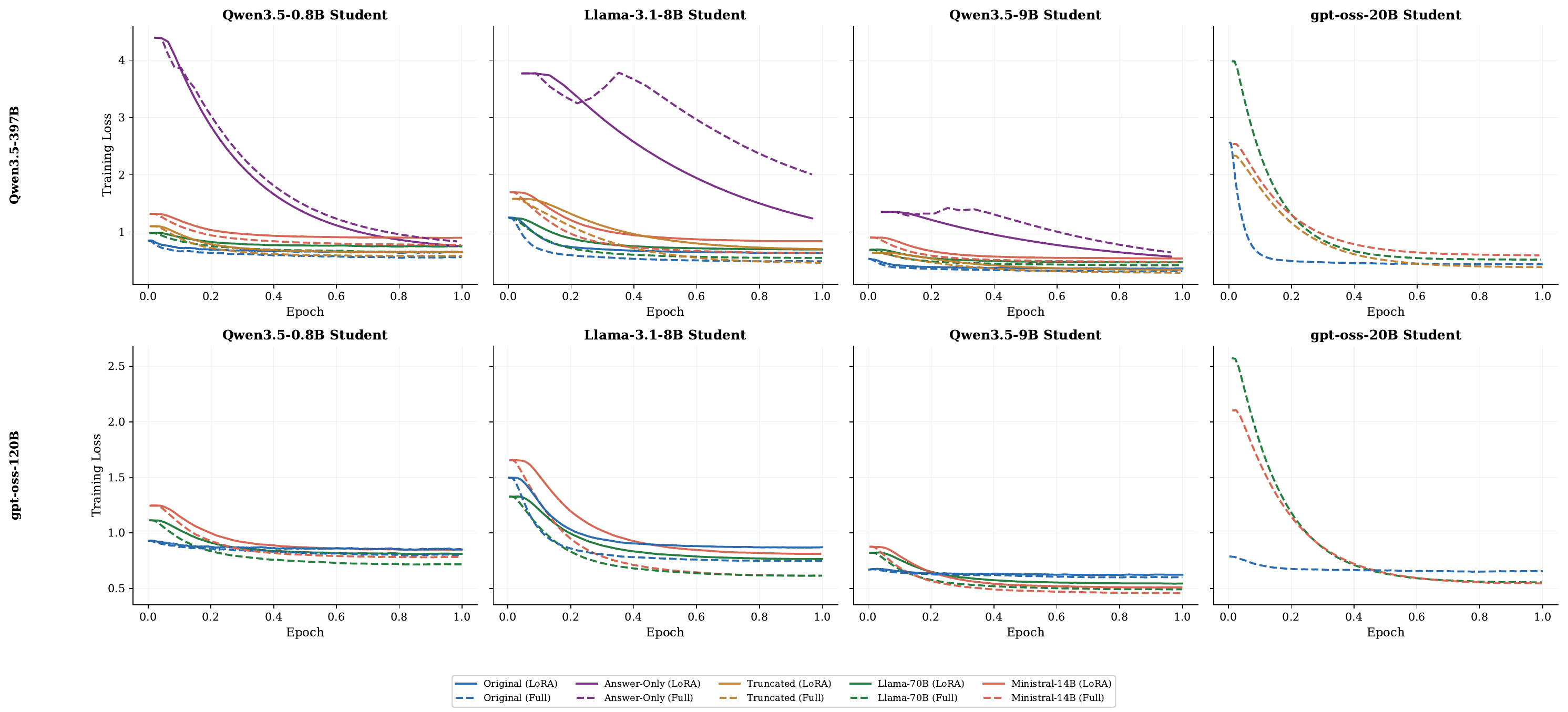}
  \caption{Training loss across the 48-run main grid plus seven Qwen-teacher truncation ablations, one row per teacher (top: Qwen3.5-397B; bottom: gpt-oss-120B), one column per student. Bold: EMA ($\alpha{=}0.08$); Solid: LoRA; dashed: full weight. gpt-oss-20B has no LoRA curve (Full only). Under the gpt-oss teacher, compressed sources usually finish below the raw curves, including at 0.8B, although the 0.8B Ministral runs remain close to raw.}
  \label{fig:loss_curves}
\end{figure*}

Within the three reasoning-trace sources, final training loss \emph{under the Qwen teacher} orders the data sources consistently in every (student, method) configuration: raw $<$ Llama-70B $<$ Ministral-14B (e.g., Qwen3.5-9B: 0.314 / 0.422 / 0.478). This matches the downstream-accuracy ordering for that teacher (Section~\ref{sec:results_eval}), so loss is a faithful ranker for these comparisons. The answer-only baseline is an outlier: its loss is sometimes lower than raw (0.289 for Qwen3.5-9B), but its downstream accuracy is far worse.

The truncated ablation is another warning that training loss alone is not supervision quality. Under the Qwen teacher, truncated targets finish with lower loss than the corresponding Ministral targets in every configuration, and in some full-fine-tuning runs even lower than raw. Yet downstream accuracy is usually below the compressed runs, especially for Llama-3.1-8B and Qwen3.5-0.8B. Matching the token budget by cutting off the raw trace therefore makes the next-token problem easier without reliably preserving the answer-bearing reasoning.

\emph{Under the gpt-oss teacher this ordering inverts}: compressed sources reach a lower final loss than raw for all larger-student configurations, although the 0.8B Ministral runs remain slightly above raw. The downstream ordering, however, still has raw $>$ compressed (Section~\ref{sec:results_eval}), so under this teacher loss is no longer a faithful ranker even within the three reasoning-trace sources.

The wall-clock benefit of compression tracks but lags the token reduction, and is smaller under the shorter-trace teacher. Under the Qwen teacher, the 0.8B student speeds up 5.8--7.6$\times$ at 12--16\% of raw tokens; the 8B/9B students gain 3.2--4.9$\times$; gpt-oss-20B sits in between (5.1--6.3$\times$). Under the gpt-oss teacher the same students gain only 2.0--4.1$\times$ at 22--30\% of raw tokens---there is simply less to remove.

% -------------------------------------------------------------------
\subsection{Downstream Evaluation}
\label{sec:results_eval}
% -------------------------------------------------------------------

\begin{figure*}[t]
  \centering
  \includegraphics[width=\textwidth]{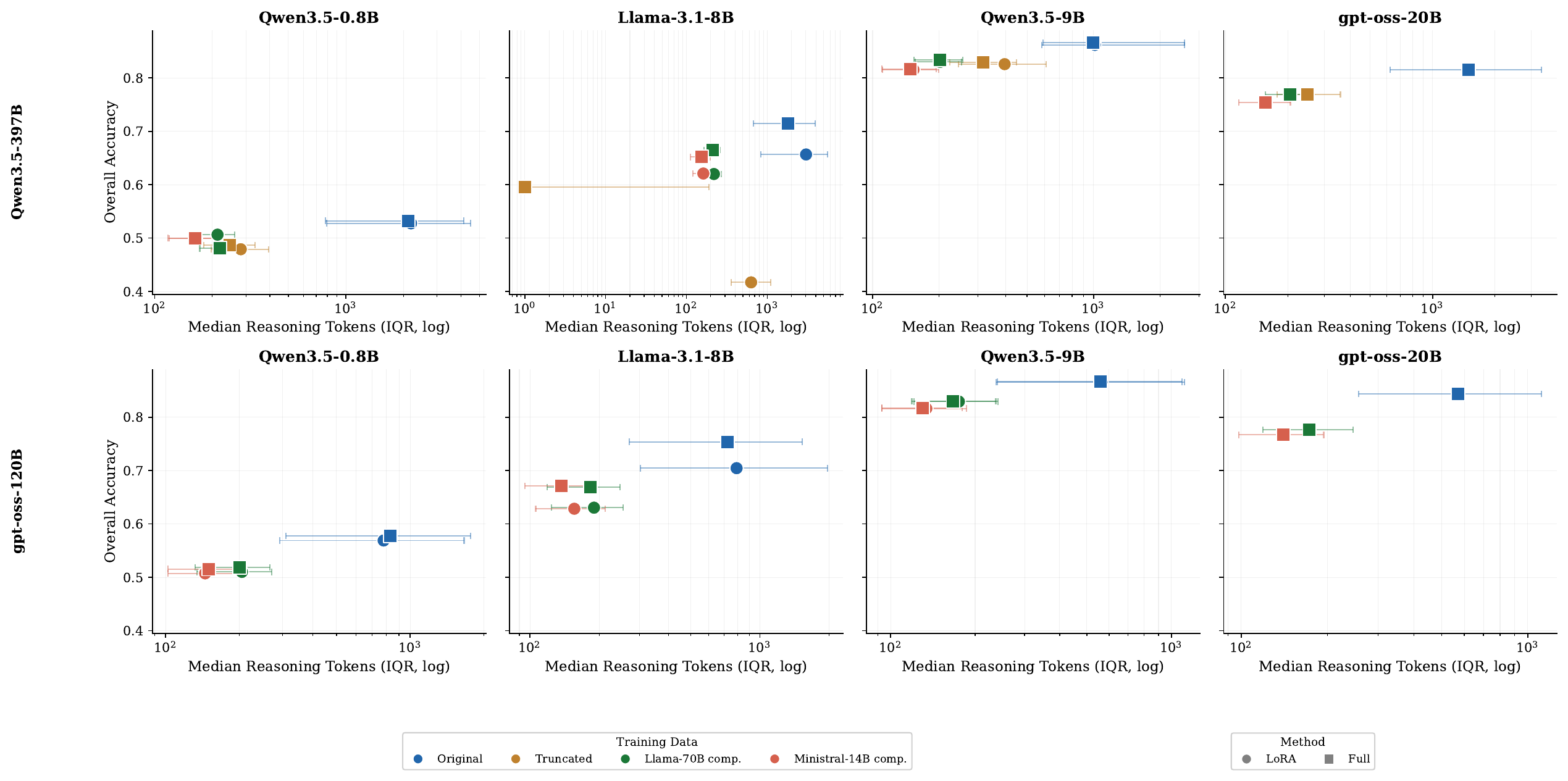}
  \caption{Per-student accuracy vs.\ median reasoning token count (log scale, IQR error bars) for the reasoning-trace runs and Qwen-teacher truncation ablations. Compression moves every student leftward (shorter); at 8B/9B/20B it also moves them slightly downward. The truncated ablation isolates token-budget reduction from model-based trace rewriting.}
  \label{fig:eval_acc_vs_len}
\end{figure*}

Table~\ref{tab:eval_results} reports overall accuracy for the main-grid reasoning-trace runs and Figure~\ref{fig:eval_acc_vs_len} plots accuracy against median reasoning tokens, including the Qwen-teacher truncation ablation. Full per-group accuracy, including answer-only rows, is in Appendix~\ref{app:eval_results} (Table~\ref{tab:eval_results_full}); matching inference statistics are in Appendix~\ref{app:eval_efficiency} (Table~\ref{tab:eval_efficiency}).

\begin{table}[t]
  \centering
  \footnotesize
  \begin{adjustbox}{max width=\columnwidth}
  \begin{tabular}{lllccc}
\toprule
\textbf{Teacher} & \textbf{Student} & \textbf{Method} & \textbf{Raw} & \textbf{L70} & \textbf{M14} \\
\midrule
\multirow{7}{*}{Qwen} & Qwen-0.8B & LoRA & \textbf{0.528} & \underline{0.506} & 0.500 \\
& Qwen-0.8B & Full & \textbf{0.532} & 0.481 & \underline{0.500} \\
& Llama-8B & LoRA & \textbf{0.657} & 0.620 & \underline{0.621} \\
& Llama-8B & Full & \textbf{0.715} & \underline{0.665} & 0.652 \\
& Qwen-9B & LoRA & \textbf{0.862} & \underline{0.830} & 0.815 \\
& Qwen-9B & Full & \textbf{0.866} & \underline{0.834} & 0.817 \\
& gpt-oss-20B & Full & \textbf{0.815} & \underline{0.769} & 0.754 \\
\midrule
\multirow{7}{*}{gpt-oss} & Qwen-0.8B & LoRA & \textbf{0.569} & \underline{0.510} & 0.507 \\
& Qwen-0.8B & Full & \textbf{0.577} & \underline{0.519} & 0.516 \\
& Llama-8B & LoRA & \textbf{0.704} & \underline{0.631} & 0.628 \\
& Llama-8B & Full & \textbf{0.754} & 0.669 & \underline{0.671} \\
& Qwen-9B & LoRA & \textbf{0.866} & \underline{0.829} & 0.816 \\
& Qwen-9B & Full & \textbf{0.866} & \underline{0.830} & 0.817 \\
& gpt-oss-20B & Full & \textbf{0.844} & \underline{0.776} & 0.767 \\
\bottomrule
\end{tabular}

  \end{adjustbox}
  \caption{Overall downstream accuracy for reasoning-trace runs. L70 and M14 are Llama-70B- and Ministral-14B-compressed traces; bold marks the within-row winner and underline marks the second best result.}
  \label{tab:eval_results}
\end{table}

\paragraph{Accuracy: raw wins at every scale, under both teachers.}
Raw teacher traces yield the highest overall accuracy in every (teacher, student, method) configuration we evaluate, including the Qwen-teacher truncation ablation (Tables~\ref{tab:eval_results} and~\ref{tab:truncated_ablation}). Under full fine-tuning, representative raw / Llama-70B / Ministral-14B rows are 0.866 / 0.834 / 0.817 for Qwen3.5-9B under the Qwen teacher and 0.844 / 0.776 / 0.767 for gpt-oss-20B under the gpt-oss teacher. The gap is largest at 8B/20B and smallest at 0.8B LoRA, where the two compressors are within $\sim$1 SE of each other. The qualitative ordering replicates across teachers.

\paragraph{Length-matched truncation: compression is not just cheaper supervision.}
The Qwen-teacher truncation ablation compares truncated raw traces against the two compressed sources (Appendix~\ref{app:truncated_ablation}). At the same training-token budget as Ministral-14B, naive truncation is substantially worse for Qwen3.5-0.8B LoRA and both Llama-3.1-8B methods. For Qwen3.5-9B and gpt-oss-20B, truncation is competitive with aggressive Ministral compression, but it still trails or ties the less aggressive Llama-70B compressor, and where it beats Ministral it emits longer inference traces (e.g., Qwen3.5-9B Full: 316 vs.\ 148 median reasoning tokens). Thus the compressed-trace result is not explained by a shorter training budget alone: rewriting the trace usually preserves useful reasoning better than cutting off the raw prefix, while raw traces remain best when accuracy is the only objective.

\paragraph{Answer-only ablation: cheap but unreliable.}
Dropping reasoning entirely is cheaper but consistently worse in our setup: its best configuration is Qwen3.5-9B LoRA at 0.781 overall (vs.\ 0.862 on raw), and LoRA at 8B/0.8B reaches 0.590/0.390 overall. Under full fine-tuning it becomes unstable: Qwen3.5-9B drops to 0.550 with 99.9\% truncation, Qwen3.5-0.8B to 0.001 with 99.9\% truncation, and Llama-3.1-8B to 0.181 with 2-token outputs. The median across the six answer-only runs falls well below the matching raw runs (see Appendix~\ref{app:eval_results}).

\paragraph{Trade-off, not domination.}
Compressed students do not Pareto-dominate raw: raw has strictly higher accuracy, compressed has strictly fewer tokens. At 9B this trades roughly 3--5 accuracy points for about 5--10$\times$ per-token efficiency, depending on teacher and compressor---worth taking when inference cost or latency dominates, not when peak accuracy does.

\paragraph{Truncation.}
Raw-trace students hit the 8{,}192-token cap on 0.6--15.4\% of records (highest at Llama-3.1-8B and 0.8B under the Qwen teacher). Compressed-trace students truncate on $\leq$1\% at the 8B/9B scales, while 20B compressed runs show a small truncation tail of 0.7--3.6\%, with gpt-oss-20B Ministral-14B a mild outlier at 3.2--3.6\%; the 0.8B Full Llama-70B-compressed run under the Qwen teacher also truncates on 23.1\% despite short medians---the small full-fine-tuned student loses control of EOT, inflating mean tokens and collapsing per-token efficiency for that run. The answer-only runs are the most extreme: Qwen3.5-9B and Qwen3.5-0.8B saturate the cap on nearly every record.

% ===================================================================
\section{Analysis}
\label{sec:analysis}
% ===================================================================

% -------------------------------------------------------------------
\subsection{Why Llama-70B Beats Ministral-14B}
\label{sec:analysis_compressors}
% -------------------------------------------------------------------

Llama-70B-compressed traces usually beat Ministral-14B-compressed on accuracy, although Ministral-14B slightly leads in Qwen-teacher Qwen3.5-0.8B Full and Llama-3.1-8B LoRA. We hypothesize that a difference in model capabilities could affect the student, as the dense 70B Llama3 model is stronger than the smaller Ministral model. We also find that Ministral-14B compresses more aggressively than Llama-70B (mean $\rho{=}0.086$ vs.\ $0.142$ under the Qwen teacher, and $0.147$ vs.\ $0.210$ under the gpt-oss teacher). This has two downstream effects. First, Ministral could merge multiple reasoning steps into single steps more aggressively, discarding intermediate computation. Second, the resulting train runs are not compute equivalent as we use the same number of epochs with a different trainable token count. Ministral-14B therefore appears to be the better choice only when brevity dominates: it has the highest Acc / 1k Tokens at Llama-3.1-8B (3.93 vs.\ 2.87) and Qwen3.5-9B LoRA (4.86 vs.\ 3.71). With Qwen3.5-9B the accuracy gap between compressors is modest, while Ministral-14B can win on Acc / 1k Tokens; on gpt-oss-20B, Llama-70B leads on Acc / 1k Tokens because Ministral-14B's 3.2--3.6\% truncation tail inflates its mean completion length.

% -------------------------------------------------------------------
\subsection{Scale Effects: Smaller Students, Larger Method Effects}
\label{sec:analysis_scale}
% -------------------------------------------------------------------

The compressed-vs-raw accuracy gap is present at every scale we evaluate, but its size depends jointly on student scale and training method. For LoRA, the gap is roughly stable across the larger students (0.862 / 0.830 / 0.815 at Qwen3.5-9B, 0.657 / 0.620 / 0.621 at Llama-3.1-8B) and narrowest at 0.528 / 0.506 / 0.500 for Qwen3.5-0.8B (raw / Llama-70B / Ministral-14B). For full fine-tuning, the gap is comparable at 9B (0.866 / 0.834 / 0.817), 8B (0.715 / 0.665 / 0.652), and 20B (0.815 / 0.769 / 0.754); at 0.8B the Full row is 0.532 / 0.481 / 0.500, with the Llama-70B-compressed run partly affected by the truncation discussed below.

Two effects plausibly combine. Smaller students have limited capacity to imitate long deliberative reasoning, so the marginal benefit of the teacher's full exploration shrinks at 0.8B, narrowing the LoRA gap. And full fine-tuning at 0.8B aggressively fits the compressed format on most examples while losing control of the EOT distribution on a tail of harder questions (the 23\% truncation in the 0.8B Llama-70B Full run); LoRA's lower-capacity adaptation does not exhibit this pattern. Compressed traces are therefore most cost-effective for LoRA at every scale, but for full fine-tuning the case is weakest at the smallest scale we test. The answer-only baseline amplifies this interaction: at Qwen3.5-9B LoRA it is genuinely competitive (0.781 vs.\ 0.862 raw), but in every other configuration it collapses, indicating that the instabilities that hurt the 0.8B compressed runs also dominate the answer-only setting once the response distribution loses its reasoning structure.

% -------------------------------------------------------------------
\subsection{Compression Transfers Brevity}
\label{sec:analysis_brevity}
% -------------------------------------------------------------------

A robust effect across all four students and both teachers is that compression transfers brevity to inference: a student trained on traces with median length $L$ produces outputs of similar median length. Under the Qwen teacher, median student reasoning is 1{,}004--1{,}085 chars (Llama-70B-compressed) and 685--745 chars (Ministral-14B-compressed) regardless of student scale; under the gpt-oss teacher these medians sit at 831--1{,}020 / 612--717 chars (Llama-70B / Ministral-14B), reflecting shorter compressed outputs in absolute terms despite milder compression ratios. Combined with $\leq$1\% truncation at the 8B/9B scales and most 20B runs, this is the property that drives per-token efficiency.

The truncated ablation shows that token-budget matching does not automatically transfer the same brevity. Even when truncated traces match Ministral's training length by construction, the resulting students often decode longer than Ministral-trained students (Qwen3.5-9B Full: 316 vs.\ 148 median reasoning tokens; gpt-oss-20B Full: 249 vs.\ 156). The learned output policy therefore depends on the structure of the shortened trace, not only its token count.

% -------------------------------------------------------------------
\subsection{Domain-Specific Accuracy Gaps}
\label{sec:analysis_domains}
% -------------------------------------------------------------------

The aggregate raw-vs-compressed gap hides a consistent per-domain structure (Appendix~\ref{app:eval_results}, Table~\ref{tab:eval_results_full}). Across the three larger students under the Qwen teacher Full, Medicine shows the largest raw advantage (4.7--8.5 points) and Science the smallest (1.4--4.0 points): medical traces carry dense factual scaffolding (drug names, dosages, contraindications) with low intrinsic redundancy, so each dropped token costs more than a token from a verbose derivation.

Math is the exception---at Llama-3.1-8B under the Qwen teacher, compression \emph{beats} raw (Full: 0.671 vs.\ 0.702 / 0.681; LoRA: 0.458 vs.\ 0.581 / 0.546). The mechanism is truncation: Llama-3.1-8B raw under the Qwen teacher truncates on 6.1--15.4\% of records (highest of any student) and these truncations concentrate on long symbolic traces, while compressed students truncate on $\leq$1\% and reach the answer. Where the raw student is comfortably below the context cap (Qwen3.5-9B, gpt-oss-20B), Math reverts to a modest raw advantage of a few points. The practical implication: prefer raw for medical-heavy workloads and any setting where the student would not otherwise truncate; compression's accuracy cost is negligible (or negative) elsewhere.

% ===================================================================
\section{Conclusion}
\label{sec:conclusion}
% ===================================================================

We studied post-hoc model-based compression of reasoning traces. Across a 48-run main grid plus seven Qwen-teacher truncation ablations, compression cuts training tokens to 12--30\% of raw, wall-clock training by 2.0--7.6$\times$, and median inference reasoning by roughly 3--19$\times$. The cost is accuracy: raw teacher traces remain best at every scale and teacher, although the gap is smallest at 0.8B LoRA. The answer-only baseline is competitive only at Qwen3.5-9B LoRA and collapses under full fine-tuning; at 0.8B Full, one Llama-compressed run also fails to terminate on 23.1\% of records. The truncation ablation shows compression is not merely a token-count artifact: naive raw-prefix truncation often loses to model-compressed traces at the same or lower inference cost, and never recovers raw-trace accuracy. The findings replicate across teachers, while the win tracks teacher verbosity. Future work should explore answer-preservation verification, domain-specific compression prompts, and a wider sweep of teachers.

\newpage
\clearpage

% ===================================================================
\section*{Limitations}
\label{sec:limitations}
% ===================================================================

\paragraph{Single, generic compression prompt.}
We use one compression prompt across all domains. A domain-specific prompt (for example, ``preserve all symbolic computation verbatim'' for mathematics) could change the per-domain accuracy gap and is not explored here.

\paragraph{Truncation.}
We use an 8{,}192-token decoding cap. Across both teachers, raw-trace students truncate on 0.6--1.8\% (Qwen3.5-9B), 1.8--15.4\% (Llama-3.1-8B), 2.5--8.6\% (Qwen3.5-0.8B), and 0.7--3.2\% (gpt-oss-20B) of records, so the reported reasoning-length advantage of compression is, at the larger scales, a slight underestimate. The 0.8B Llama-70B Full compressed run under the Qwen teacher additionally truncates on 23.1\% of records despite its short median outputs, which suppresses its per-token efficiency relative to the LoRA counterpart; we discuss this anomaly in Section~\ref{sec:results_eval}. The answer-only runs are the most extreme: Qwen3.5-9B and Qwen3.5-0.8B saturate the cap on essentially every record. These failure modes indicate that the base model may require a different training setup to learn zero-shot answering.

% ===================================================================
\section*{Acknowledgements}
\label{sec:acknowledgements}
% ===================================================================

Generative AI models such as ChatGPT and Claude were used to assist with coding and editing of the manuscript.

% ===================================================================
% References
% ===================================================================

\bibliography{main}

\begin{thebibliography}{43}
\providecommand{\natexlab}[1]{#1}

\bibitem[{Aggarwal and Welleck(2025)}]{aggarwal2025l}
Pranjal Aggarwal and Sean Welleck. 2025.
\newblock \href {https://openreview.net/forum?id=4jdIxXBNve} {L1: Controlling how long a reasoning model thinks with reinforcement learning}.
\newblock In \emph{Second Conference on Language Modeling}.

\bibitem[{Alonso et~al.(2024)Alonso, Oronoz, and Agerri}]{Alonso_2024}
Iñigo Alonso, Maite Oronoz, and Rodrigo Agerri. 2024.
\newblock \href {https://doi.org/10.1016/j.artmed.2024.102938} {Medexpqa: Multilingual benchmarking of large language models for medical question answering}.
\newblock \emph{Artificial Intelligence in Medicine}, 155:102938.

\bibitem[{Aytes et~al.(2025)Aytes, Baek, and Hwang}]{aytes2025sketch}
Simon~A. Aytes, Jinheon Baek, and Sung~Ju Hwang. 2025.
\newblock \href {https://doi.org/10.18653/v1/2025.emnlp-main.1236} {Sketch-of-thought: Efficient {LLM} reasoning with adaptive cognitive-inspired sketching}.
\newblock In \emph{Proceedings of the 2025 Conference on Empirical Methods in Natural Language Processing}, pages 24296--24320, Suzhou, China. Association for Computational Linguistics.

\bibitem[{Bisk et~al.(2020)Bisk, Zellers, Bras, Gao, and Choi}]{bisk_piqa_2020}
Yonatan Bisk, Rowan Zellers, Ronan~Le Bras, Jianfeng Gao, and Yejin Choi. 2020.
\newblock {{PIQA}}: {{Reasoning}} about {{Physical Commonsense}} in {{Natural Language}}.
\newblock In \emph{Thirty-{{Fourth AAAI Conference}} on {{Artificial Intelligence}}}.

\bibitem[{Chen et~al.(2025)Chen, Xu, Liang, He, Pang, Yu, Song, Liu, Zhou, Zhang, Wang, Tu, Mi, and Yu}]{chen2025do}
Xingyu Chen, Jiahao Xu, Tian Liang, Zhiwei He, Jianhui Pang, Dian Yu, Linfeng Song, Qiuzhi Liu, Mengfei Zhou, Zhuosheng Zhang, Rui Wang, Zhaopeng Tu, Haitao Mi, and Dong Yu. 2025.
\newblock \href {https://openreview.net/forum?id=MSbU3L7V00} {Do {NOT} think that much for 2+3=? on the overthinking of long reasoning models}.
\newblock In \emph{Forty-second International Conference on Machine Learning}.

\bibitem[{Clark et~al.(2019)Clark, Lee, Chang, Kwiatkowski, Collins, and Toutanova}]{clark_boolq_2019}
Christopher Clark, Kenton Lee, Ming-Wei Chang, Tom Kwiatkowski, Michael Collins, and Kristina Toutanova. 2019.
\newblock \href {https://doi.org/10.18653/v1/N19-1300} {{{BoolQ}}: {{Exploring}} the {{Surprising Difficulty}} of {{Natural Yes}}/{{No Questions}}}.
\newblock In \emph{Proceedings of the 2019 {{Conference}} of the {{North American Chapter}} of the {{Association}} for {{Computational Linguistics}}: {{Human Language Technologies}}, {{Volume}} 1 ({{Long}} and {{Short Papers}})}, pages 2924--2936, Minneapolis, Minnesota. Association for Computational Linguistics.

\bibitem[{Clark et~al.(2018)Clark, Cowhey, Etzioni, Khot, Sabharwal, Schoenick, and Tafjord}]{clark2018thinksolvedquestionanswering}
Peter Clark, Isaac Cowhey, Oren Etzioni, Tushar Khot, Ashish Sabharwal, Carissa Schoenick, and Oyvind Tafjord. 2018.
\newblock \href {https://arxiv.org/abs/1803.05457} {Think you have solved question answering? try arc, the ai2 reasoning challenge}.
\newblock \emph{Preprint}, arXiv:1803.05457.

\bibitem[{Cobbe et~al.(2021)Cobbe, Kosaraju, Bavarian, Chen, Jun, Kaiser, Plappert, Tworek, Hilton, Nakano, Hesse, and Schulman}]{cobbe2021trainingverifierssolvemath}
Karl Cobbe, Vineet Kosaraju, Mohammad Bavarian, Mark Chen, Heewoo Jun, Lukasz Kaiser, Matthias Plappert, Jerry Tworek, Jacob Hilton, Reiichiro Nakano, Christopher Hesse, and John Schulman. 2021.
\newblock \href {https://arxiv.org/abs/2110.14168} {Training verifiers to solve math word problems}.
\newblock \emph{Preprint}, arXiv:2110.14168.

\bibitem[{Dao(2024)}]{dao2024flashattention}
Tri Dao. 2024.
\newblock \href {https://openreview.net/forum?id=mZn2Xyh9Ec} {Flashattention-2: Faster attention with better parallelism and work partitioning}.
\newblock In \emph{The Twelfth International Conference on Learning Representations}.

\bibitem[{Dubey et~al.(2024)Dubey, Jauhri, Pandey, Kadian, {Al-Dahle}, Letman, Mathur, Schelten, Yang, Fan, Goyal, Hartshorn, Yang, Mitra, Sravankumar, Korenev, Hinsvark, Rao, Zhang, Rodriguez, Gregerson, Spataru, Roziere, Biron, Tang, Chern, Caucheteux, Nayak, Bi, Marra, McConnell, Keller, Touret, Wu, Wong, Ferrer, Nikolaidis, Allonsius, Song, Pintz, Livshits, Esiobu, Choudhary, Mahajan, {Garcia-Olano}, Perino, Hupkes, Lakomkin, AlBadawy, Lobanova, Dinan, Smith, Radenovic, Zhang, Synnaeve, Lee, Anderson, Nail, Mialon, Pang, Cucurell, Nguyen, Korevaar, Xu, Touvron, Zarov, Ibarra, Kloumann, Misra, Evtimov, Copet, Lee, Geffert, Vranes, Park, Mahadeokar, Shah, van~der Linde, Billock, Hong, Lee, Fu, Chi, Huang, Liu, Wang, Yu, Bitton, Spisak, Park, Rocca, Johnstun, Saxe, Jia, Alwala, Upasani, Plawiak, Li, Heafield, Stone, {El-Arini}, Iyer, Malik, Chiu, Bhalla, {Rantala-Yeary}, van~der Maaten, Chen, Tan, Jenkins, Martin, Madaan, Malo, Blecher, Landzaat, de~Oliveira, Muzzi, Pasupuleti, Singh, Paluri, Kardas, Oldham,
  Rita, Pavlova, Kambadur, Lewis, Si, Singh, Hassan, Goyal, Torabi, Bashlykov, Bogoychev, Chatterji, Duchenne, {\c C}elebi, Alrassy, Zhang, Li, Vasic, Weng, Bhargava, Dubal, Krishnan, Koura, Xu, He, Dong, Srinivasan, Ganapathy, Calderer, Cabral, Stojnic, Raileanu, Girdhar, Patel, Sauvestre, Polidoro, Sumbaly, Taylor, Silva, Hou, Wang, Hosseini, Chennabasappa, Singh, Bell, Kim, Edunov, Nie, Narang, Raparthy, Shen, Wan, Bhosale, Zhang, Vandenhende, Batra, Whitman, Sootla, Collot, Gururangan, Borodinsky, Herman, Fowler, Sheasha, Georgiou, Scialom, Speckbacher, Mihaylov, Xiao, Karn, Goswami, Gupta, Ramanathan, Kerkez, Gonguet, Do, Vogeti, Petrovic, Chu, Xiong, Fu, Meers, Martinet, Wang, Tan, Xie, Jia, Wang, Goldschlag, Gaur, Babaei, Wen, Song, Zhang, Li, Mao, Coudert, Yan, Chen, Papakipos, Singh, Grattafiori, Jain, Kelsey, Shajnfeld, Gangidi, Victoria, Goldstand, Menon, Sharma, Boesenberg, Vaughan, Baevski, Feinstein, Kallet, Sangani, Yunus, Lupu, Alvarado, Caples, Gu, Ho, Poulton, Ryan, Ramchandani, Franco,
  Saraf, Chowdhury, Gabriel, Bharambe, Eisenman, Yazdan, James, Maurer, Leonhardi, Huang, Loyd, Paola, Paranjape, Liu, Wu, Ni, Hancock, Wasti, Spence, Stojkovic, Gamido, Montalvo, Parker, Burton, Mejia, Wang, Kim, Zhou, Hu, Chu, Cai, Tindal, Feichtenhofer, Civin, Beaty, Kreymer, Li, Wyatt, Adkins, Xu, Testuggine, David, Parikh, Liskovich, Foss, Wang, Le, Holland, Dowling, Jamil, Montgomery, Presani, Hahn, Wood, Brinkman, Arcaute, Dunbar, Smothers, Sun, Kreuk, Tian, Ozgenel, Caggioni, Guzm{\'a}n, Kanayet, Seide, Florez, Schwarz, Badeer, Swee, Halpern, Thattai, Herman, Sizov, Guangyi, Zhang, Lakshminarayanan, Shojanazeri, Zou, Wang, Zha, Habeeb, Rudolph, Suk, Aspegren, Goldman, Damlaj, Molybog, Tufanov, Veliche, Gat, Weissman, Geboski, Kohli, Asher, Gaya, Marcus, Tang, Chan, Zhen, Reizenstein, Teboul, Zhong, Jin, Yang, Cummings, Carvill, Shepard, McPhie, Torres, Ginsburg, Wang, Wu, U, Saxena, Prasad, Khandelwal, Zand, Matosich, Veeraraghavan, Michelena, Li, Huang, Chawla, Lakhotia, Huang, Chen, Garg, A, Silva,
  Bell, Zhang, Guo, Yu, Moshkovich, Wehrstedt, Khabsa, Avalani, Bhatt, Tsimpoukelli, Mankus, Hasson, Lennie, Reso, Groshev, Naumov, Lathi, Keneally, Seltzer, Valko, Restrepo, Patel, Vyatskov, Samvelyan, Clark, Macey, Wang, Hermoso, Metanat, Rastegari, Bansal, Santhanam, Parks, White, Bawa, Singhal, Egebo, Usunier, Laptev, Dong, Zhang, Cheng, Chernoguz, Hart, Salpekar, Kalinli, Kent, Parekh, Saab, Balaji, Rittner, Bontrager, Roux, Dollar, Zvyagina, Ratanchandani, Yuvraj, Liang, Alao, Rodriguez, Ayub, Murthy, Nayani, Mitra, Li, Hogan, Battey, Wang, Maheswari, Howes, Rinott, Bondu, Datta, Chugh, Hunt, Dhillon, Sidorov, Pan, Verma, Yamamoto, Ramaswamy, Lindsay, Lindsay, Feng, Lin, Zha, Shankar, Zhang, Zhang, Wang, Agarwal, Sajuyigbe, Chintala, Max, Chen, Kehoe, Satterfield, Govindaprasad, Gupta, Cho, Virk, Subramanian, Choudhury, Goldman, Remez, Glaser, Best, Kohler, Robinson, Li, Zhang, Matthews, Chou, Shaked, Vontimitta, Ajayi, Montanez, Mohan, Kumar, Mangla, Albiero, Ionescu, Poenaru, Mihailescu, Ivanov, Li,
  Wang, Jiang, Bouaziz, Constable, Tang, Wang, Wu, Wang, Xia, Wu, Gao, Chen, Hu, Jia, Qi, Li, Zhang, Zhang, Adi, Nam, Yu, Wang, Hao, Qian, He, Rait, DeVito, Rosnbrick, Wen, Yang, and Zhao}]{grattafiori2024llama3}
Abhimanyu Dubey, Abhinav Jauhri, Abhinav Pandey, Abhishek Kadian, Ahmad {Al-Dahle}, Aiesha Letman, Akhil Mathur, Alan Schelten, Amy Yang, Angela Fan, Anirudh Goyal, Anthony Hartshorn, Aobo Yang, Archi Mitra, Archie Sravankumar, Artem Korenev, Arthur Hinsvark, Arun Rao, Aston Zhang, and 516 others. 2024.
\newblock \href {https://doi.org/10.48550/arXiv.2407.21783} {The {{Llama}} 3 {{Herd}} of {{Models}}}.
\newblock \emph{Preprint}, arXiv:2407.21783.

\bibitem[{Guo et~al.(2025)Guo, Yang, Zhang, Song, Wang, Zhu, Xu, Zhang, Ma, Bi, Zhang, Yu, Wu, Wu, Gou, Shao, Li, Gao, Liu, Xue, Wang, Wu, Feng, Lu, Zhao, Deng, Ruan, Dai, Chen, Ji, Li, Lin, Dai, Luo, Hao, Chen, Li, Zhang, Xu, Ding, Gao, Qu, Li, Guo, Li, Chen, Yuan, Tu, Qiu, Li, Cai, Ni, Liang, Chen, Dong, Hu, You, Gao, Guan, Huang, Yu, Wang, Zhang, Zhao, Wang, Zhang, Xu, Xia, Zhang, Zhang, Tang, Zhou, Li, Wang, Li, Tian, Huang, Zhang, Wang, Chen, Du, Ge, Zhang, Pan, Wang, Chen, Jin, Chen, Lu, Zhou, Chen, Ye, Wang, Yu, Zhou, Pan, Li, Zhou, Wu, Yun, Pei, Sun, Wang, Zeng, Liu, Liang, Gao, Yu, Zhang, Xiao, An, Liu, Wang, Chen, Nie, Cheng, Liu, Xie, Liu, Yang, Li, Su, Lin, Li, Jin, Shen, Chen, Sun, Wang, Song, Zhou, Wang, Shan, Li, Wang, Wei, Zhang, Xu, Li, Zhao, Sun, Wang, Yu, Zhang, Shi, Xiong, He, Piao, Wang, Tan, Ma, Liu, Guo, Ou, Wang, Gong, Zou, He, Xiong, Luo, You, Liu, Zhou, Zhu, Huang, Li, Zheng, Zhu, Ma, Tang, Zha, Yan, Ren, Ren, Sha, Fu, Xu, Xie, Zhang, Hao, Ma, Yan, Wu, Gu, Zhu, Liu, Li, Xie, Song,
  Pan, Huang, Xu, Zhang, and Zhang}]{deepseek2025r1}
Daya Guo, Dejian Yang, Haowei Zhang, Junxiao Song, Peiyi Wang, Qihao Zhu, Runxin Xu, Ruoyu Zhang, Shirong Ma, Xiao Bi, Xiaokang Zhang, Xingkai Yu, Yu~Wu, Z.~F. Wu, Zhibin Gou, Zhihong Shao, Zhuoshu Li, Ziyi Gao, Aixin Liu, and 175 others. 2025.
\newblock \href {https://doi.org/10.1038/s41586-025-09422-z} {{{DeepSeek-R1}} incentivizes reasoning in {{LLMs}} through reinforcement learning}.
\newblock \emph{Nature}, 645(8081):633--638.

\bibitem[{Han et~al.(2025)Han, Wang, Fang, Zhao, Ma, and Chen}]{han2024token}
Tingxu Han, Zhenting Wang, Chunrong Fang, Shiyu Zhao, Shiqing Ma, and Zhenyu Chen. 2025.
\newblock \href {https://doi.org/10.18653/v1/2025.findings-acl.1274} {Token-budget-aware {LLM} reasoning}.
\newblock In \emph{Findings of the Association for Computational Linguistics: ACL 2025}, pages 24842--24855, Vienna, Austria. Association for Computational Linguistics.

\bibitem[{Hendrycks et~al.(2021)Hendrycks, Burns, Basart, Zou, Mazeika, Song, and Steinhardt}]{hendrycks2021measuring}
Dan Hendrycks, Collin Burns, Steven Basart, Andy Zou, Mantas Mazeika, Dawn Song, and Jacob Steinhardt. 2021.
\newblock \href {https://openreview.net/forum?id=d7KBjmI3GmQ} {Measuring massive multitask language understanding}.
\newblock In \emph{International Conference on Learning Representations}.

\bibitem[{Ho et~al.(2023)Ho, Schmid, and Yun}]{ho2023large}
Namgyu Ho, Laura Schmid, and Se-Young Yun. 2023.
\newblock \href {https://doi.org/10.18653/v1/2023.acl-long.830} {Large language models are reasoning teachers}.
\newblock In \emph{Proceedings of the 61st Annual Meeting of the Association for Computational Linguistics (Volume 1: Long Papers)}, pages 14852--14882, Toronto, Canada. Association for Computational Linguistics.

\bibitem[{Hu et~al.(2022)Hu, yelong shen, Wallis, Allen-Zhu, Li, Wang, Wang, and Chen}]{hu2022lora}
Edward~J Hu, yelong shen, Phillip Wallis, Zeyuan Allen-Zhu, Yuanzhi Li, Shean Wang, Lu~Wang, and Weizhu Chen. 2022.
\newblock \href {https://openreview.net/forum?id=nZeVKeeFYf9} {Lo{RA}: Low-rank adaptation of large language models}.
\newblock In \emph{International Conference on Learning Representations}.

\bibitem[{Jin et~al.(2021)Jin, Pan, Oufattole, Weng, Fang, and Szolovits}]{jin_what_2021}
Di~Jin, Eileen Pan, Nassim Oufattole, Wei-Hung Weng, Hanyi Fang, and Peter Szolovits. 2021.
\newblock \href {https://doi.org/10.3390/app11146421} {What {{Disease Does This Patient Have}}? {{A Large-Scale Open Domain Question Answering Dataset}} from {{Medical Exams}}}.
\newblock \emph{Applied Sciences}, 11(14).

\bibitem[{Kang et~al.(2025)Kang, Sun, Chen, and Zou}]{kang2025c3ot}
Yu~Kang, Xianghui Sun, Liangyu Chen, and Wei Zou. 2025.
\newblock \href {https://doi.org/10.1609/aaai.v39i23.34608} {C3ot: generating shorter chain-of-thought without compromising effectiveness}.
\newblock In \emph{Proceedings of the Thirty-Ninth AAAI Conference on Artificial Intelligence and Thirty-Seventh Conference on Innovative Applications of Artificial Intelligence and Fifteenth Symposium on Educational Advances in Artificial Intelligence}, AAAI'25/IAAI'25/EAAI'25. AAAI Press.

\bibitem[{Kojima et~al.(2022)Kojima, Gu, Reid, Matsuo, and Iwasawa}]{kojima2022large}
Takeshi Kojima, Shixiang~Shane Gu, Machel Reid, Yutaka Matsuo, and Yusuke Iwasawa. 2022.
\newblock Large language models are zero-shot reasoners.
\newblock In \emph{Proceedings of the 36th International Conference on Neural Information Processing Systems}, NIPS '22, Red Hook, NY, USA. Curran Associates Inc.

\bibitem[{Kwon et~al.(2023)Kwon, Li, Zhuang, Sheng, Zheng, Yu, Gonzalez, Zhang, and Stoica}]{kwon_efficient_2023}
Woosuk Kwon, Zhuohan Li, Siyuan Zhuang, Ying Sheng, Lianmin Zheng, Cody~Hao Yu, Joseph Gonzalez, Hao Zhang, and Ion Stoica. 2023.
\newblock \href {https://doi.org/10.1145/3600006.3613165} {Efficient {{Memory Management}} for {{Large Language Model Serving}} with {{PagedAttention}}}.
\newblock In \emph{Proceedings of the 29th {{Symposium}} on {{Operating Systems Principles}}}, {{SOSP}} '23, pages 611--626, New York, NY, USA. Association for Computing Machinery.

\bibitem[{Lin et~al.(2022)Lin, Hilton, and Evans}]{lin-etal-2022-truthfulqa}
Stephanie Lin, Jacob Hilton, and Owain Evans. 2022.
\newblock \href {https://doi.org/10.18653/v1/2022.acl-long.229} {{T}ruthful{QA}: Measuring how models mimic human falsehoods}.
\newblock In \emph{Proceedings of the 60th Annual Meeting of the Association for Computational Linguistics (Volume 1: Long Papers)}, pages 3214--3252, Dublin, Ireland. Association for Computational Linguistics.

\bibitem[{Ling et~al.(2017)Ling, Yogatama, Dyer, and Blunsom}]{ling2017programinductionrationalegeneration}
Wang Ling, Dani Yogatama, Chris Dyer, and Phil Blunsom. 2017.
\newblock \href {https://arxiv.org/abs/1705.04146} {Program induction by rationale generation : Learning to solve and explain algebraic word problems}.
\newblock \emph{Preprint}, arXiv:1705.04146.

\bibitem[{Liu et~al.(2026)Liu, Khandelwal, Subramanian, Jouault, Rastogi, Sadé, Jeffares, Jiang, Cahill, Gavaudan, Sablayrolles, Héliou, You, Ehrenberg, Lo, Eliseev, Calvi, Sooriyarachchi, Bout, Rozière, Monicault, Lanfranchi, Barreau, Courtot, Grattarola, Dabert, de~las Casas, Chane-Sane, Ahmed, Berrada, Ecrepont, Guinet, Novikov, Kunsch, Lample, Martin, Gupta, Ludziejewski, Rute, Studnia, Amar, Delas, Roberts, Yadav, Chandu, Jain, Aitchison, Fainsin, Blier, Zhao, Martin, Saulnier, Gao, Buyl, Jennings, Pellat, Prins, Poirée, Guillaumin, Dinot, Futeral, Darrin, Augustin, Chiquier, Schimpf, Grinsztajn, Gupta, Raghuraman, Bousquet, Duchenne, Wang, von Platen, Jacob, Wambergue, Kurylowicz, Muddireddy, Chagniot, Stock, Agrawal, Torroba, Sauvestre, Soletskyi, Menneer, Vaze, Barry, Gandhi, Waghjale, Gandhi, Ghosh, Mishra, Aithal, Antoniak, Scao, Cachet, Sorg, Lavril, Saada, Chabal, Foubert, Robert, Wang, Lawson, Bewley, Bewley, Edwards, Jamil, Tomasini, Nemychnikova, Phung, Maladière, Richard, Bouaziz, Li,
  Marshall, Li, Yang, Ouahidi, Wang, Tang, and Ramzi}]{liu2026ministral3}
Alexander~H. Liu, Kartik Khandelwal, Sandeep Subramanian, Victor Jouault, Abhinav Rastogi, Adrien Sadé, Alan Jeffares, Albert Jiang, Alexandre Cahill, Alexandre Gavaudan, Alexandre Sablayrolles, Amélie Héliou, Amos You, Andy Ehrenberg, Andy Lo, Anton Eliseev, Antonia Calvi, Avinash Sooriyarachchi, Baptiste Bout, and 101 others. 2026.
\newblock \href {https://arxiv.org/abs/2601.08584} {Ministral 3}.
\newblock \emph{Preprint}, arXiv:2601.08584.

\bibitem[{Luo et~al.(2025)Luo, Shen, He, Wang, Liu, Li, Tan, Cao, and Tao}]{luo2025o1}
Haotian Luo, Li~Shen, Haiying He, Yibo Wang, Shiwei Liu, Wei Li, Naiqiang Tan, Xiaochun Cao, and Dacheng Tao. 2025.
\newblock \href {https://doi.org/10.48550/arXiv.2501.12570} {O1-pruner: Length-harmonizing fine-tuning for o1-like reasoning pruning}.
\newblock \emph{CoRR}, abs/2501.12570.

\bibitem[{Magister et~al.(2023)Magister, Mallinson, Adamek, Malmi, and Severyn}]{magister2023teaching}
Lucie~Charlotte Magister, Jonathan Mallinson, Jakub Adamek, Eric Malmi, and Aliaksei Severyn. 2023.
\newblock \href {https://arxiv.org/abs/2212.08410} {Teaching small language models to reason}.
\newblock \emph{Preprint}, arXiv:2212.08410.

\bibitem[{Munkhbat et~al.(2025)Munkhbat, Ho, Kim, Yang, Kim, and Yun}]{munkhbat2025self}
Tergel Munkhbat, Namgyu Ho, Seo~Hyun Kim, Yongjin Yang, Yujin Kim, and Se-Young Yun. 2025.
\newblock \href {https://doi.org/10.18653/v1/2025.findings-acl.1289} {Self-training elicits concise reasoning in large language models}.
\newblock In \emph{Findings of the Association for Computational Linguistics: ACL 2025}, pages 25127--25152, Vienna, Austria. Association for Computational Linguistics.

\bibitem[{OpenAI et~al.(2025)OpenAI, :, Agarwal, Ahmad, Ai, Altman, Applebaum, Arbus, Arora, Bai, Baker, Bao, Barak, Bennett, Bertao, Brett, Brevdo, Brockman, Bubeck, Chang, Chen, Chen, Cheung, Clark, Cook, Dukhan, Dvorak, Fives, Fomenko, Garipov, Georgiev, Glaese, Gogineni, Goucher, Gross, Guzman, Hallman, Hehir, Heidecke, Helyar, Hu, Huet, Huh, Jain, Johnson, Koch, Kofman, Kundel, Kwon, Kyrylov, Le, Leclerc, Lennon, Lessans, Lezcano-Casado, Li, Li, Lin, Liss, Lily, Liu, Liu, Lu, Lu, Martinovic, McCallum, McGrath, McKinney, McLaughlin, Mei, Mostovoy, Mu, Myles, Neitz, Nichol, Pachocki, Paino, Palmie, Pantuliano, Parascandolo, Park, Pathak, Paz, Peran, Pimenov, Pokrass, Proehl, Qiu, Raila, Raso, Ren, Richardson, Robinson, Rotsted, Salman, Sanjeev, Schwarzer, Sculley, Sikchi, Simon, Singhal, Song, Stuckey, Sun, Tillet, Toizer, Tsimpourlas, Vyas, Wallace, Wang, Wang, Watkins, Weil, Wendling, Whinnery, Whitney, Wong, Yang, Yang, Yasunaga, Ying, Zaremba, Zhan, Zhang, Zhang, Zhang, and
  Zhao}]{openai2025gptoss120bgptoss20bmodel}
OpenAI, :, Sandhini Agarwal, Lama Ahmad, Jason Ai, Sam Altman, Andy Applebaum, Edwin Arbus, Rahul~K. Arora, Yu~Bai, Bowen Baker, Haiming Bao, Boaz Barak, Ally Bennett, Tyler Bertao, Nivedita Brett, Eugene Brevdo, Greg Brockman, Sebastien Bubeck, and 108 others. 2025.
\newblock \href {https://arxiv.org/abs/2508.10925} {gpt-oss-120b \& gpt-oss-20b model card}.
\newblock \emph{Preprint}, arXiv:2508.10925.

\bibitem[{Pal et~al.(2022)Pal, Umapathi, and Sankarasubbu}]{pal_medmcqa_2022}
Ankit Pal, Logesh~Kumar Umapathi, and Malaikannan Sankarasubbu. 2022.
\newblock {{MedMCQA}}: {{A Large-scale Multi-Subject Multi-Choice Dataset}} for {{Medical}} domain {{Question Answering}}.
\newblock In \emph{Proceedings of the {{Conference}} on {{Health}}, {{Inference}}, and {{Learning}}}, volume 174 of \emph{Proceedings of {{Machine Learning Research}}}, pages 248--260. PMLR.

\bibitem[{Patel et~al.(2021)Patel, Bhattamishra, and Goyal}]{patel-etal-2021-nlp}
Arkil Patel, Satwik Bhattamishra, and Navin Goyal. 2021.
\newblock \href {https://doi.org/10.18653/v1/2021.naacl-main.168} {Are {NLP} models really able to solve simple math word problems?}
\newblock In \emph{Proceedings of the 2021 Conference of the North American Chapter of the Association for Computational Linguistics: Human Language Technologies}, pages 2080--2094, Online. Association for Computational Linguistics.

\bibitem[{Rein et~al.(2024)Rein, Hou, Stickland, Petty, Pang, Dirani, Michael, and Bowman}]{rein2024gpqa}
David Rein, Betty~Li Hou, Asa~Cooper Stickland, Jackson Petty, Richard~Yuanzhe Pang, Julien Dirani, Julian Michael, and Samuel~R. Bowman. 2024.
\newblock \href {https://openreview.net/forum?id=Ti67584b98} {{GPQA}: A graduate-level google-proof q\&a benchmark}.
\newblock In \emph{First Conference on Language Modeling}.

\bibitem[{Roy and Roth(2015)}]{roy-roth-2015-solving}
Subhro Roy and Dan Roth. 2015.
\newblock \href {https://doi.org/10.18653/v1/D15-1202} {Solving general arithmetic word problems}.
\newblock In \emph{Proceedings of the 2015 Conference on Empirical Methods in Natural Language Processing}, pages 1743--1752, Lisbon, Portugal. Association for Computational Linguistics.

\bibitem[{Sakaguchi et~al.(2021)Sakaguchi, Bras, Bhagavatula, and Choi}]{sakaguchi_winogrande_2021}
Keisuke Sakaguchi, Ronan~Le Bras, Chandra Bhagavatula, and Yejin Choi. 2021.
\newblock \href {https://doi.org/10.1145/3474381} {{{WinoGrande}}: {{An Adversarial Winograd Schema Challenge}} at {{Scale}}}.
\newblock \emph{Commun. ACM}, 64(9):99--106.

\bibitem[{Shen et~al.(2025)Shen, Zhang, Huang, Shi, Zhang, Yan, Wang, Wang, Liu, and Lian}]{shen2025dast}
Yi~Shen, Jian Zhang, Jieyun Huang, Shuming Shi, Wenjing Zhang, Jiangze Yan, Ning Wang, Kai Wang, Zhaoxiang Liu, and Shiguo Lian. 2025.
\newblock \href {https://doi.org/10.18653/v1/2025.emnlp-industry.160} {{DAST}: Difficulty-adaptive slow-thinking for large reasoning models}.
\newblock In \emph{Proceedings of the 2025 Conference on Empirical Methods in Natural Language Processing: Industry Track}, pages 2322--2331, Suzhou (China). Association for Computational Linguistics.

\bibitem[{Shridhar et~al.(2023)Shridhar, Stolfo, and Sachan}]{shridhar2023distilling}
Kumar Shridhar, Alessandro Stolfo, and Mrinmaya Sachan. 2023.
\newblock \href {https://doi.org/10.18653/v1/2023.findings-acl.441} {Distilling reasoning capabilities into smaller language models}.
\newblock In \emph{Findings of the Association for Computational Linguistics: ACL 2023}, pages 7059--7073, Toronto, Canada. Association for Computational Linguistics.

\bibitem[{Talmor et~al.(2019)Talmor, Herzig, Lourie, and Berant}]{talmor-etal-2019-commonsenseqa}
Alon Talmor, Jonathan Herzig, Nicholas Lourie, and Jonathan Berant. 2019.
\newblock \href {https://doi.org/10.18653/v1/N19-1421} {{C}ommonsense{QA}: A question answering challenge targeting commonsense knowledge}.
\newblock In \emph{Proceedings of the 2019 Conference of the North {A}merican Chapter of the Association for Computational Linguistics: Human Language Technologies, Volume 1 (Long and Short Papers)}, pages 4149--4158, Minneapolis, Minnesota. Association for Computational Linguistics.

\bibitem[{Team(2025)}]{qwen2025qwq}
Qwen Team. 2025.
\newblock {{QwQ-32B}}: {{Embracing}} the {{Power}} of {{Reinforcement Learning}}.
\newblock https://qwen.ai/blog?id=qwq-32b.

\bibitem[{Wei et~al.(2022)Wei, Wang, Schuurmans, Bosma, Ichter, Xia, Chi, Le, and Zhou}]{wei2022chain}
Jason Wei, Xuezhi Wang, Dale Schuurmans, Maarten Bosma, Brian Ichter, Fei Xia, Ed~H. Chi, Quoc~V. Le, and Denny Zhou. 2022.
\newblock Chain-of-thought prompting elicits reasoning in large language models.
\newblock In \emph{Proceedings of the 36th International Conference on Neural Information Processing Systems}, NIPS '22, Red Hook, NY, USA. Curran Associates Inc.

\bibitem[{Wijmans et~al.(2025)Wijmans, Huval, Hertzberg, Koltun, and Kraehenbuehl}]{wijmans2025cut}
Erik Wijmans, Brody Huval, Alexander Hertzberg, Vladlen Koltun, and Philipp Kraehenbuehl. 2025.
\newblock \href {https://openreview.net/forum?id=E4Fk3YuG56} {Cut your losses in large-vocabulary language models}.
\newblock In \emph{The Thirteenth International Conference on Learning Representations}.

\bibitem[{Xia et~al.(2025)Xia, Leong, Wang, Li, and Li}]{xia2025tokenskip}
Heming Xia, Chak~Tou Leong, Wenjie Wang, Yongqi Li, and Wenjie Li. 2025.
\newblock \href {https://doi.org/10.18653/v1/2025.emnlp-main.165} {{T}oken{S}kip: Controllable chain-of-thought compression in {LLM}s}.
\newblock In \emph{Proceedings of the 2025 Conference on Empirical Methods in Natural Language Processing}, pages 3351--3363, Suzhou, China. Association for Computational Linguistics.

\bibitem[{Xu et~al.(2025)Xu, Xie, Zhao, and He}]{xu2025chain}
Silei Xu, Wenhao Xie, Lingxiao Zhao, and Pengcheng He. 2025.
\newblock \href {https://arxiv.org/abs/2502.18600} {Chain of draft: Thinking faster by writing less}.
\newblock \emph{Preprint}, arXiv:2502.18600.

\bibitem[{Yang et~al.(2025{\natexlab{a}})Yang, Li, Yang, Zhang, Hui, Zheng, Yu, Gao, Huang, Lv, Zheng, Liu, Zhou, Huang, Hu, Ge, Wei, Lin, Tang, Yang, Tu, Zhang, Yang, Yang, Zhou, Zhou, Lin, Dang, Bao, Yang, Yu, Deng, Li, Xue, Li, Zhang, Wang, Zhu, Men, Gao, Liu, Luo, Li, Tang, Yin, Ren, Wang, Zhang, Ren, Fan, Su, Zhang, Zhang, Wan, Liu, Wang, Cui, Zhang, Zhou, and Qiu}]{qwen2025qwen3}
An~Yang, Anfeng Li, Baosong Yang, Beichen Zhang, Binyuan Hui, Bo~Zheng, Bowen Yu, Chang Gao, Chengen Huang, Chenxu Lv, Chujie Zheng, Dayiheng Liu, Fan Zhou, Fei Huang, Feng Hu, Hao Ge, Haoran Wei, Huan Lin, Jialong Tang, and 41 others. 2025{\natexlab{a}}.
\newblock Qwen3 {{Technical Report}}.
\newblock https://arxiv.org/abs/2505.09388v1.

\bibitem[{Yang et~al.(2025{\natexlab{b}})Yang, Lin, and Yu}]{yang2025think}
Junjie Yang, Ke~Lin, and Xing Yu. 2025{\natexlab{b}}.
\newblock \href {https://arxiv.org/abs/2504.03234} {Think when you need: Self-adaptive chain-of-thought learning}.
\newblock \emph{Preprint}, arXiv:2504.03234.

\bibitem[{Zellers et~al.(2019)Zellers, Holtzman, Bisk, Farhadi, and Choi}]{zellers_hellaswag_2019}
Rowan Zellers, Ari Holtzman, Yonatan Bisk, Ali Farhadi, and Yejin Choi. 2019.
\newblock \href {https://doi.org/10.18653/v1/P19-1472} {{{HellaSwag}}: {{Can}} a {{Machine Really Finish Your Sentence}}?}
\newblock In \emph{Proceedings of the 57th {{Annual Meeting}} of the {{Association}} for {{Computational Linguistics}}}, pages 4791--4800, Florence, Italy. Association for Computational Linguistics.

\bibitem[{Zhang et~al.(2026)Zhang, Yu, Pan, Jin, Fu, Cai, Lin, and Ye}]{zhang2026tokensqueeze}
Yuxiang Zhang, Zhengxu Yu, Weihang Pan, Zhongming Jin, Qiang Fu, Deng Cai, Binbin Lin, and Jieping Ye. 2026.
\newblock \href {https://openreview.net/forum?id=Wc1VZ2bVJn} {Tokensqueeze: Performance-preserving compression for reasoning {LLM}s}.
\newblock In \emph{The Thirty-ninth Annual Conference on Neural Information Processing Systems}.

\end{thebibliography}

\FloatBarrier

\appendix
\onecolumn

% ===================================================================
\section{Per-Dataset Trace Generation}
\label{app:trace_generation}
% ===================================================================

Table~\ref{tab:trace_generation} reports per-dataset trace-generation statistics for both teachers under rejection sampling.

\begin{table*}[h]
  \centering
  \begin{adjustbox}{max width=\textwidth}
\begin{tabular}{lllrrr}
\toprule
\textbf{Generator} & \textbf{Domain} & \textbf{Dataset} & \textbf{Total} & \textbf{Avg.\ Reason (chars)} & \textbf{Avg.\ Total (tok)} \\
\midrule
\multirow{13}{*}{Qwen3.5-397B} & Math & AQUA-RAT & 87,441 & 7,583 & 3,427 \\
 & Math & GSM8k & 7,186 & 4,688 & 1,689 \\
 & Math & MultiArith & 416 & 1,750 & 604 \\
 & Math & SVAMP & 960 & 2,926 & 974 \\
 & Science & ARC-Challenge & 1,079 & 4,264 & 1,196 \\
 & Science & GPQA Diamond & 135 & 18,734 & 6,923 \\
 & Logic & CommonsenseQA & 8,593 & 6,418 & 1,778 \\
 & Medical & MedExpQA & 395 & 11,541 & 3,569 \\
 & Medical & MedMCQA & 167,277 & 6,801 & 2,015 \\
 & Medical & MedQA & 9,778 & 10,806 & 3,490 \\
 & MMLU & Mmlu-Medical & 33 & 9,876 & 2,810 \\
 & MMLU & Mmlu-Stem & 42 & 9,347 & 3,412 \\
 & \multicolumn{2}{l}{\textbf{Total}} & \textbf{283,335} & \textbf{--} & \textbf{---} \\
\midrule
\multirow{13}{*}{gpt-oss-120B} & Math & AQUA-RAT & 89,796 & 3,360 & 1,166 \\
 & Math & GSM8k & 7,268 & 1,878 & 633 \\
 & Math & MultiArith & 413 & 716 & 316 \\
 & Math & SVAMP & 967 & 1,368 & 466 \\
 & Science & ARC-Challenge & 1,079 & 1,554 & 477 \\
 & Science & GPQA Diamond & 152 & 13,407 & 4,191 \\
 & Logic & CommonsenseQA & 8,434 & 4,292 & 1,138 \\
 & Medical & MedExpQA & 395 & 5,321 & 1,432 \\
 & Medical & MedMCQA & 163,480 & 4,043 & 1,072 \\
 & Medical & MedQA & 9,850 & 4,537 & 1,304 \\
 & MMLU & Mmlu-Medical & 33 & 3,286 & 888 \\
 & MMLU & Mmlu-Stem & 44 & 4,031 & 1,253 \\
 & \multicolumn{2}{l}{\textbf{Total}} & \textbf{281,911} & \textbf{--} & \textbf{---} \\
\bottomrule
\end{tabular}
\end{adjustbox}

  \caption{Trace generation results. Total is the number of correct traces retained. Avg.\ Reason and Avg.\ Total report mean reasoning trace length (characters) and total completion tokens per correct trace.}
  \label{tab:trace_generation}
\end{table*}

\FloatBarrier

% ===================================================================
\section{Per-Dataset Compression}
\label{app:compression}
% ===================================================================

Table~\ref{tab:compression_full} reports compression statistics by teacher, compressor, and dataset.

\begin{table*}[h]
  \centering
  \begin{adjustbox}{max width=\textwidth}
\begin{tabular}{lll rr c rr c}
\toprule
& & & \multicolumn{3}{c}{\textbf{Llama-3.3-70B-Instruct}} & \multicolumn{3}{c}{\textbf{Ministral-3-14B-Instruct}} \\
\cmidrule(lr){4-6}
\cmidrule(lr){7-9}
\textbf{Generator} & \textbf{Domain} & \textbf{Dataset} & \textbf{Avg.\ Comp.} & \textbf{Comp.\ Tok} & \textbf{$\rho$} & \textbf{Avg.\ Comp.} & \textbf{Comp.\ Tok} & \textbf{$\rho$} \\
& &  & \textbf{(chars)} & &  & \textbf{(chars)} & & \\
\midrule
\multirow{11}{*}{Qwen3.5-397B} & \multirow{4}{*}{Math} & AQUA-RAT     & 1,034 & 340 & 0.136 & 478 & 185 & \textbf{0.063} \\
 &  & GSM8k        & 612 & 172 & 0.131 & 296 & 117 & \textbf{0.063} \\
 &  & MultiArith   & 443 & 109 & 0.253 & 221 & 80 & \textbf{0.126} \\
 &  & SVAMP        & 522 & 127 & 0.178 & 294 & 93 & \textbf{0.101} \\
\cmidrule(lr){2-9}
 & \multirow{2}{*}{Science} & ARC-Challenge & 911 & 185 & 0.214 & 522 & 113 & \textbf{0.122} \\
 &  & GPQA Diamond & 1,658 & 479 & 0.089 & 1,020 & 315 & \textbf{0.054} \\
\cmidrule(lr){2-9}
 & \multirow{1}{*}{Logic} & CommonsenseQA & 862 & 181 & 0.134 & 564 & 124 & \textbf{0.088} \\
\cmidrule(lr){2-9}
 & \multirow{3}{*}{Medical} & MedExpQA     & 1,317 & 286 & 0.114 & 1,113 & 266 & \textbf{0.096} \\
 &  & MedMCQA      & 1,002 & 222 & 0.147 & 677 & 158 & \textbf{0.100} \\
 &  & MedQA        & 1,369 & 300 & 0.127 & 1,044 & 247 & \textbf{0.097} \\
\cmidrule(lr){2-9}
 & \multicolumn{2}{l}{\textbf{Overall}} & \textbf{1,008} & \textbf{---} & \textbf{0.142} & \textbf{613} & \textbf{---} & \textbf{\underline{0.086}} \\
\midrule
\multirow{11}{*}{gpt-oss-120B} & \multirow{4}{*}{Math} & AQUA-RAT     & 722 & 216 & 0.215 & 438 & 170 & \textbf{0.130} \\
 &  & GSM8k        & 505 & 138 & 0.269 & 263 & 98 & \textbf{0.140} \\
 &  & MultiArith   & 332 & 85 & 0.464 & 160 & 55 & \textbf{0.224} \\
 &  & SVAMP        & 418 & 101 & 0.306 & 239 & 72 & \textbf{0.175} \\
\cmidrule(lr){2-9}
 & \multirow{2}{*}{Science} & ARC-Challenge & 624 & 124 & 0.401 & 439 & 95 & \textbf{0.282} \\
 &  & GPQA Diamond & 1,446 & 403 & 0.108 & 1,012 & 317 & \textbf{0.076} \\
\cmidrule(lr){2-9}
 & \multirow{1}{*}{Logic} & CommonsenseQA & 706 & 148 & 0.164 & 538 & 121 & \textbf{0.125} \\
\cmidrule(lr){2-9}
 & \multirow{3}{*}{Medical} & MedExpQA     & 1,185 & 253 & 0.223 & 1,051 & 249 & \textbf{0.198} \\
 &  & MedMCQA      & 833 & 181 & 0.206 & 610 & 141 & \textbf{0.151} \\
 &  & MedQA        & 1,094 & 233 & 0.241 & 954 & 225 & \textbf{0.210} \\
\cmidrule(lr){2-9}
 & \multicolumn{2}{l}{\textbf{Overall}} & \textbf{793} & \textbf{---} & \textbf{0.210} & \textbf{554} & \textbf{---} & \textbf{\underline{0.147}} \\
\bottomrule
\end{tabular}
\end{adjustbox}

  \caption{Compression statistics by (teacher, compressor, dataset). $\rho$ is the ratio of mean compressed length to mean original length. The two halves of the table report the same compressors applied to the Qwen3.5-397B teacher's traces (top) and the gpt-oss-120B teacher's traces (bottom); ratios are uniformly milder under the shorter-trace gpt-oss teacher.}
  \label{tab:compression_full}
\end{table*}

\FloatBarrier

% ===================================================================
\section{Training-Time Efficiency}
\label{app:training_efficiency}
% ===================================================================

Table~\ref{tab:efficiency} reports wall-clock training time and source-dataset trainable tokens for each student, method, teacher, and data source.

\begin{table*}[h]
  \centering
  \begin{adjustbox}{max width=\textwidth}
  \begin{tabular}{lll rrrrr rrrrr}
\toprule
& & & \multicolumn{5}{c}{\textbf{Training Time (h)}} & \multicolumn{5}{c}{\textbf{Training Tokens (M)}} \\
\cmidrule(lr){4-8} \cmidrule(lr){9-13}
\textbf{Generator} & \textbf{Student} & \textbf{Method}  & Original & Answer-Only & Truncated & Llama-70B & Ministral-14B  & Original & Answer-Only & Truncated & Llama-70B & Ministral-14B \\
\midrule
  \multirow{7}{*}{Qwen3.5-397B} & \multirow{2}{*}{Qwen3.5-0.8B} & LoRA & 1.1 & 0.1\;{\scriptsize(16.8$\times$)} & 0.1\;{\scriptsize(7.8$\times$)} & 0.2\;{\scriptsize(5.9$\times$)} & 0.1\;{\scriptsize(7.6$\times$)} & 635 & 25 & 75 & 102 & 74 \\
   &  & Full & 1.1 & 0.1\;{\scriptsize(15.8$\times$)} & 0.1\;{\scriptsize(7.6$\times$)} & 0.2\;{\scriptsize(5.8$\times$)} & 0.1\;{\scriptsize(7.3$\times$)} & 635 & 25 & 75 & 102 & 74 \\
\cmidrule(lr){2-13}
   & \multirow{2}{*}{Llama-3.1-8B} & LoRA & 2.7 & 0.1\;{\scriptsize(28.0$\times$)} & 0.3\;{\scriptsize(9.4$\times$)} & 0.8\;{\scriptsize(3.5$\times$)} & 0.6\;{\scriptsize(4.8$\times$)} & 589 & 23 & 71 & 97 & 71 \\
   &  & Full & 3.1 & 0.1\;{\scriptsize(24.8$\times$)} & 0.3\;{\scriptsize(9.1$\times$)} & 0.9\;{\scriptsize(3.6$\times$)} & 0.6\;{\scriptsize(4.9$\times$)} & 589 & 23 & 71 & 97 & 71 \\
\cmidrule(lr){2-13}
   & \multirow{2}{*}{Qwen3.5-9B} & LoRA & 2.7 & 0.1\;{\scriptsize(22.0$\times$)} & 0.3\;{\scriptsize(8.2$\times$)} & 0.8\;{\scriptsize(3.2$\times$)} & 0.6\;{\scriptsize(4.4$\times$)} & 635 & 24 & 74 & 102 & 74 \\
   &  & Full & 3.3 & 0.2\;{\scriptsize(19.2$\times$)} & 0.4\;{\scriptsize(8.0$\times$)} & 1.0\;{\scriptsize(3.3$\times$)} & 0.8\;{\scriptsize(4.4$\times$)} & 635 & 24 & 74 & 102 & 74 \\
\cmidrule(lr){2-13}
   & \multirow{1}{*}{gpt-oss-20B} & Full & 3.1 & --- & 0.5\;{\scriptsize(6.4$\times$)} & 0.6\;{\scriptsize(5.1$\times$)} & 0.5\;{\scriptsize(6.3$\times$)} & 600 & --- & 88 & 113 & 87 \\
\midrule
  \multirow{7}{*}{gpt-oss-120B} & \multirow{2}{*}{Qwen3.5-0.8B} & LoRA & 0.6 & --- & --- & 0.2\;{\scriptsize(3.6$\times$)} & 0.1\;{\scriptsize(4.1$\times$)} & 315 & --- & --- & 81 & 69 \\
   &  & Full & 0.5 & --- & --- & 0.2\;{\scriptsize(3.5$\times$)} & 0.1\;{\scriptsize(3.9$\times$)} & 315 & --- & --- & 81 & 69 \\
\cmidrule(lr){2-13}
   & \multirow{2}{*}{Llama-3.1-8B} & LoRA & 1.4 & --- & --- & 0.6\;{\scriptsize(2.2$\times$)} & 0.5\;{\scriptsize(2.6$\times$)} & 298 & --- & --- & 78 & 66 \\
   &  & Full & 1.5 & --- & --- & 0.7\;{\scriptsize(2.3$\times$)} & 0.6\;{\scriptsize(2.6$\times$)} & 298 & --- & --- & 78 & 66 \\
\cmidrule(lr){2-13}
   & \multirow{2}{*}{Qwen3.5-9B} & LoRA & 1.3 & --- & --- & 0.7\;{\scriptsize(2.0$\times$)} & 0.6\;{\scriptsize(2.3$\times$)} & 315 & --- & --- & 81 & 69 \\
   &  & Full & 1.7 & --- & --- & 0.8\;{\scriptsize(2.0$\times$)} & 0.7\;{\scriptsize(2.3$\times$)} & 315 & --- & --- & 81 & 69 \\
\cmidrule(lr){2-13}
   & \multirow{1}{*}{gpt-oss-20B} & Full & 1.6 & --- & --- & 0.5\;{\scriptsize(3.2$\times$)} & 0.5\;{\scriptsize(3.5$\times$)} & 310 & --- & --- & 94 & 83 \\
\bottomrule
\end{tabular}

  \end{adjustbox}
  \caption{Training-time efficiency; speedup vs.\ raw in parentheses. Across both teachers, compression cuts wall-clock by 3.5--7.6$\times$ for 0.8B, 2.0--4.9$\times$ for 8B/9B, and 3.2--6.3$\times$ for 20B. The answer-only baseline is cheaper (15.8--28.0$\times$) but is uniformly worse on downstream accuracy (Section~\ref{sec:results_eval}).}
  \label{tab:efficiency}
\end{table*}

\FloatBarrier

% ===================================================================
\section{Downstream Evaluation Results}
\label{app:eval_results}
% ===================================================================

Table~\ref{tab:eval_results_full} reports per-group and overall downstream accuracy for the 48 main-grid runs plus seven Qwen-teacher truncation ablations.

\begin{table*}[h]
  \centering
  \begin{adjustbox}{max width=\textwidth}
\begin{tabular}{llllccccccc}
\toprule
\multirow{2}{*}{\textbf{Generator}} & \multirow{2}{*}{\textbf{Student}} & \multirow{2}{*}{\textbf{Data}} & \multirow{2}{*}{\textbf{Method}} & \multicolumn{4}{c}{\textbf{In-distribution}} & \multicolumn{2}{c}{\textbf{OOD}} & \multirow{2}{*}{\textbf{Overall}} \\
\cmidrule(lr){5-8} \cmidrule(lr){9-10}
& & & & Math & Sci. & Med. & Cmsn. & Reason. & Know. & \\
\midrule
\multirow{34}{*}{Qwen3.5-397B} & \multirow{10}{*}{Qwen3.5-0.8B} & \multirow{2}{*}{Raw traces} & LoRA & \textbf{0.596}{\scriptsize\,$\pm$0.023} & \textbf{0.725}{\scriptsize\,$\pm$0.026} & \textbf{0.405}{\scriptsize\,$\pm$0.013} & \textbf{0.599}{\scriptsize\,$\pm$0.027} & \underline{0.547}{\scriptsize\,$\pm$0.027} & \textbf{0.555}{\scriptsize\,$\pm$0.026} & \textbf{0.528}{\scriptsize\,$\pm$0.005} \\
 &  &  & Full & \textbf{0.641}{\scriptsize\,$\pm$0.023} & \textbf{0.711}{\scriptsize\,$\pm$0.026} & \textbf{0.437}{\scriptsize\,$\pm$0.013} & \textbf{0.653}{\scriptsize\,$\pm$0.027} & \textbf{0.549}{\scriptsize\,$\pm$0.027} & \textbf{0.565}{\scriptsize\,$\pm$0.026} & \textbf{0.532}{\scriptsize\,$\pm$0.005} \\
\cmidrule(lr){3-11}
 &  & \multirow{2}{*}{Answer-Only} & LoRA & 0.075{\scriptsize\,$\pm$0.012} & 0.645{\scriptsize\,$\pm$0.027} & 0.368{\scriptsize\,$\pm$0.013} & 0.559{\scriptsize\,$\pm$0.028} & 0.443{\scriptsize\,$\pm$0.027} & 0.476{\scriptsize\,$\pm$0.026} & 0.390{\scriptsize\,$\pm$0.007} \\
 &  &  & Full & 0.017{\scriptsize\,$\pm$0.006} & 0.000{\scriptsize\,$\pm$0.002} & 0.000{\scriptsize\,$\pm$0.000} & 0.000{\scriptsize\,$\pm$0.002} & 0.000{\scriptsize\,$\pm$0.002} & 0.000{\scriptsize\,$\pm$0.001} & 0.001{\scriptsize\,$\pm$0.000} \\
\cmidrule(lr){3-11}
 &  & \multirow{2}{*}{Truncated} & LoRA & 0.123{\scriptsize\,$\pm$0.015} & 0.688{\scriptsize\,$\pm$0.027} & 0.377{\scriptsize\,$\pm$0.013} & 0.563{\scriptsize\,$\pm$0.028} & 0.459{\scriptsize\,$\pm$0.027} & 0.518{\scriptsize\,$\pm$0.026} & 0.479{\scriptsize\,$\pm$0.005} \\
 &  &  & Full & 0.120{\scriptsize\,$\pm$0.015} & 0.671{\scriptsize\,$\pm$0.027} & 0.392{\scriptsize\,$\pm$0.013} & \underline{0.622}{\scriptsize\,$\pm$0.027} & 0.466{\scriptsize\,$\pm$0.027} & \underline{0.536}{\scriptsize\,$\pm$0.026} & 0.487{\scriptsize\,$\pm$0.005} \\
\cmidrule(lr){3-11}
 &  & \multirow{2}{*}{Llama-70B comp.} & LoRA & \underline{0.513}{\scriptsize\,$\pm$0.023} & \underline{0.711}{\scriptsize\,$\pm$0.026} & \underline{0.394}{\scriptsize\,$\pm$0.013} & \underline{0.567}{\scriptsize\,$\pm$0.028} & \textbf{0.549}{\scriptsize\,$\pm$0.027} & \underline{0.534}{\scriptsize\,$\pm$0.026} & \underline{0.506}{\scriptsize\,$\pm$0.005} \\
 &  &  & Full & 0.499{\scriptsize\,$\pm$0.023} & 0.638{\scriptsize\,$\pm$0.028} & 0.395{\scriptsize\,$\pm$0.013} & 0.587{\scriptsize\,$\pm$0.028} & 0.520{\scriptsize\,$\pm$0.027} & 0.468{\scriptsize\,$\pm$0.026} & 0.481{\scriptsize\,$\pm$0.005} \\
\cmidrule(lr){3-11}
 &  & \multirow{2}{*}{Ministral-14B comp.} & LoRA & 0.491{\scriptsize\,$\pm$0.023} & 0.693{\scriptsize\,$\pm$0.026} & 0.385{\scriptsize\,$\pm$0.013} & 0.558{\scriptsize\,$\pm$0.028} & 0.505{\scriptsize\,$\pm$0.027} & 0.529{\scriptsize\,$\pm$0.026} & 0.500{\scriptsize\,$\pm$0.005} \\
 &  &  & Full & \underline{0.526}{\scriptsize\,$\pm$0.023} & \underline{0.686}{\scriptsize\,$\pm$0.027} & \underline{0.409}{\scriptsize\,$\pm$0.013} & 0.575{\scriptsize\,$\pm$0.028} & \underline{0.527}{\scriptsize\,$\pm$0.027} & 0.522{\scriptsize\,$\pm$0.026} & \underline{0.500}{\scriptsize\,$\pm$0.005} \\
\cmidrule(lr){2-11}
 & \multirow{10}{*}{Llama-3.1-8B} & \multirow{2}{*}{Raw traces} & LoRA & 0.458{\scriptsize\,$\pm$0.023} & \textbf{0.801}{\scriptsize\,$\pm$0.023} & \textbf{0.600}{\scriptsize\,$\pm$0.013} & \textbf{0.722}{\scriptsize\,$\pm$0.025} & \textbf{0.565}{\scriptsize\,$\pm$0.027} & \textbf{0.701}{\scriptsize\,$\pm$0.024} & \textbf{0.657}{\scriptsize\,$\pm$0.005} \\
 &  &  & Full & 0.671{\scriptsize\,$\pm$0.022} & \textbf{0.834}{\scriptsize\,$\pm$0.021} & \textbf{0.671}{\scriptsize\,$\pm$0.012} & \textbf{0.810}{\scriptsize\,$\pm$0.022} & \textbf{0.627}{\scriptsize\,$\pm$0.027} & \textbf{0.750}{\scriptsize\,$\pm$0.023} & \textbf{0.715}{\scriptsize\,$\pm$0.004} \\
\cmidrule(lr){3-11}
 &  & \multirow{2}{*}{Answer-Only} & LoRA & 0.131{\scriptsize\,$\pm$0.016} & 0.703{\scriptsize\,$\pm$0.026} & 0.533{\scriptsize\,$\pm$0.013} & 0.672{\scriptsize\,$\pm$0.026} & 0.514{\scriptsize\,$\pm$0.027} & 0.661{\scriptsize\,$\pm$0.025} & 0.590{\scriptsize\,$\pm$0.005} \\
 &  &  & Full & 0.040{\scriptsize\,$\pm$0.009} & 0.196{\scriptsize\,$\pm$0.023} & 0.211{\scriptsize\,$\pm$0.011} & 0.191{\scriptsize\,$\pm$0.022} & 0.199{\scriptsize\,$\pm$0.022} & 0.205{\scriptsize\,$\pm$0.021} & 0.181{\scriptsize\,$\pm$0.004} \\
\cmidrule(lr){3-11}
 &  & \multirow{2}{*}{Truncated} & LoRA & 0.071{\scriptsize\,$\pm$0.012} & 0.459{\scriptsize\,$\pm$0.029} & 0.389{\scriptsize\,$\pm$0.013} & 0.369{\scriptsize\,$\pm$0.027} & 0.370{\scriptsize\,$\pm$0.026} & 0.450{\scriptsize\,$\pm$0.026} & 0.417{\scriptsize\,$\pm$0.005} \\
 &  &  & Full & 0.377{\scriptsize\,$\pm$0.023} & 0.755{\scriptsize\,$\pm$0.025} & 0.569{\scriptsize\,$\pm$0.013} & 0.758{\scriptsize\,$\pm$0.024} & 0.515{\scriptsize\,$\pm$0.027} & 0.655{\scriptsize\,$\pm$0.025} & 0.596{\scriptsize\,$\pm$0.005} \\
\cmidrule(lr){3-11}
 &  & \multirow{2}{*}{Llama-70B comp.} & LoRA & \textbf{0.581}{\scriptsize\,$\pm$0.023} & \underline{0.779}{\scriptsize\,$\pm$0.024} & \underline{0.562}{\scriptsize\,$\pm$0.013} & \underline{0.704}{\scriptsize\,$\pm$0.026} & \underline{0.560}{\scriptsize\,$\pm$0.027} & \underline{0.672}{\scriptsize\,$\pm$0.024} & 0.620{\scriptsize\,$\pm$0.005} \\
 &  &  & Full & \textbf{0.702}{\scriptsize\,$\pm$0.021} & \underline{0.803}{\scriptsize\,$\pm$0.023} & \underline{0.611}{\scriptsize\,$\pm$0.013} & \underline{0.779}{\scriptsize\,$\pm$0.023} & \underline{0.594}{\scriptsize\,$\pm$0.027} & 0.704{\scriptsize\,$\pm$0.024} & \underline{0.665}{\scriptsize\,$\pm$0.005} \\
\cmidrule(lr){3-11}
 &  & \multirow{2}{*}{Ministral-14B comp.} & LoRA & \underline{0.546}{\scriptsize\,$\pm$0.023} & 0.762{\scriptsize\,$\pm$0.024} & 0.544{\scriptsize\,$\pm$0.013} & 0.684{\scriptsize\,$\pm$0.026} & 0.542{\scriptsize\,$\pm$0.027} & 0.658{\scriptsize\,$\pm$0.025} & \underline{0.621}{\scriptsize\,$\pm$0.005} \\
 &  &  & Full & \underline{0.681}{\scriptsize\,$\pm$0.022} & 0.794{\scriptsize\,$\pm$0.023} & 0.599{\scriptsize\,$\pm$0.013} & 0.765{\scriptsize\,$\pm$0.024} & 0.568{\scriptsize\,$\pm$0.027} & \underline{0.706}{\scriptsize\,$\pm$0.024} & 0.652{\scriptsize\,$\pm$0.005} \\
\cmidrule(lr){2-11}
 & \multirow{10}{*}{Qwen3.5-9B} & \multirow{2}{*}{Raw traces} & LoRA & \textbf{0.943}{\scriptsize\,$\pm$0.011} & \textbf{0.945}{\scriptsize\,$\pm$0.013} & \textbf{0.755}{\scriptsize\,$\pm$0.011} & \textbf{0.848}{\scriptsize\,$\pm$0.020} & \textbf{0.922}{\scriptsize\,$\pm$0.015} & \textbf{0.854}{\scriptsize\,$\pm$0.018} & \textbf{0.862}{\scriptsize\,$\pm$0.003} \\
 &  &  & Full & \textbf{0.954}{\scriptsize\,$\pm$0.010} & \textbf{0.951}{\scriptsize\,$\pm$0.012} & \textbf{0.770}{\scriptsize\,$\pm$0.011} & \textbf{0.854}{\scriptsize\,$\pm$0.020} & \textbf{0.930}{\scriptsize\,$\pm$0.014} & \textbf{0.852}{\scriptsize\,$\pm$0.018} & \textbf{0.866}{\scriptsize\,$\pm$0.003} \\
\cmidrule(lr){3-11}
 &  & \multirow{2}{*}{Answer-Only} & LoRA & 0.319{\scriptsize\,$\pm$0.022} & 0.911{\scriptsize\,$\pm$0.016} & 0.653{\scriptsize\,$\pm$0.012} & 0.798{\scriptsize\,$\pm$0.023} & 0.765{\scriptsize\,$\pm$0.023} & 0.799{\scriptsize\,$\pm$0.021} & 0.781{\scriptsize\,$\pm$0.004} \\
 &  &  & Full & 0.369{\scriptsize\,$\pm$0.023} & 0.754{\scriptsize\,$\pm$0.025} & 0.556{\scriptsize\,$\pm$0.013} & 0.673{\scriptsize\,$\pm$0.026} & 0.585{\scriptsize\,$\pm$0.027} & 0.651{\scriptsize\,$\pm$0.025} & 0.550{\scriptsize\,$\pm$0.005} \\
\cmidrule(lr){3-11}
 &  & \multirow{2}{*}{Truncated} & LoRA & 0.612{\scriptsize\,$\pm$0.023} & \underline{0.938}{\scriptsize\,$\pm$0.014} & \underline{0.718}{\scriptsize\,$\pm$0.012} & \underline{0.819}{\scriptsize\,$\pm$0.022} & 0.828{\scriptsize\,$\pm$0.021} & \underline{0.830}{\scriptsize\,$\pm$0.020} & 0.826{\scriptsize\,$\pm$0.004} \\
 &  &  & Full & 0.615{\scriptsize\,$\pm$0.023} & \underline{0.939}{\scriptsize\,$\pm$0.014} & \underline{0.724}{\scriptsize\,$\pm$0.012} & 0.827{\scriptsize\,$\pm$0.021} & 0.841{\scriptsize\,$\pm$0.020} & \underline{0.834}{\scriptsize\,$\pm$0.019} & 0.830{\scriptsize\,$\pm$0.004} \\
\cmidrule(lr){3-11}
 &  & \multirow{2}{*}{Llama-70B comp.} & LoRA & \underline{0.904}{\scriptsize\,$\pm$0.014} & 0.928{\scriptsize\,$\pm$0.015} & 0.715{\scriptsize\,$\pm$0.012} & 0.817{\scriptsize\,$\pm$0.022} & 0.869{\scriptsize\,$\pm$0.019} & 0.826{\scriptsize\,$\pm$0.020} & \underline{0.830}{\scriptsize\,$\pm$0.004} \\
 &  &  & Full & \underline{0.921}{\scriptsize\,$\pm$0.013} & 0.937{\scriptsize\,$\pm$0.014} & 0.723{\scriptsize\,$\pm$0.012} & \underline{0.851}{\scriptsize\,$\pm$0.020} & \underline{0.873}{\scriptsize\,$\pm$0.018} & 0.826{\scriptsize\,$\pm$0.020} & \underline{0.834}{\scriptsize\,$\pm$0.004} \\
\cmidrule(lr){3-11}
 &  & \multirow{2}{*}{Ministral-14B comp.} & LoRA & 0.889{\scriptsize\,$\pm$0.015} & 0.932{\scriptsize\,$\pm$0.014} & 0.695{\scriptsize\,$\pm$0.012} & 0.810{\scriptsize\,$\pm$0.022} & \underline{0.870}{\scriptsize\,$\pm$0.018} & 0.816{\scriptsize\,$\pm$0.020} & 0.815{\scriptsize\,$\pm$0.004} \\
 &  &  & Full & 0.895{\scriptsize\,$\pm$0.014} & 0.928{\scriptsize\,$\pm$0.015} & 0.711{\scriptsize\,$\pm$0.012} & 0.840{\scriptsize\,$\pm$0.021} & 0.860{\scriptsize\,$\pm$0.019} & 0.819{\scriptsize\,$\pm$0.020} & 0.817{\scriptsize\,$\pm$0.004} \\
\cmidrule(lr){2-11}
 & \multirow{4}{*}{gpt-oss-20B} & Raw traces & Full & \textbf{0.946}{\scriptsize\,$\pm$0.011} & \textbf{0.927}{\scriptsize\,$\pm$0.015} & \textbf{0.718}{\scriptsize\,$\pm$0.012} & \textbf{0.829}{\scriptsize\,$\pm$0.021} & \textbf{0.888}{\scriptsize\,$\pm$0.017} & \textbf{0.826}{\scriptsize\,$\pm$0.020} & \textbf{0.815}{\scriptsize\,$\pm$0.004} \\
\cmidrule(lr){3-11}
 &  & Truncated & Full & 0.453{\scriptsize\,$\pm$0.023} & \underline{0.905}{\scriptsize\,$\pm$0.017} & \underline{0.671}{\scriptsize\,$\pm$0.012} & 0.792{\scriptsize\,$\pm$0.023} & 0.776{\scriptsize\,$\pm$0.023} & 0.788{\scriptsize\,$\pm$0.021} & \underline{0.769}{\scriptsize\,$\pm$0.004} \\
\cmidrule(lr){3-11}
 &  & Llama-70B comp. & Full & \underline{0.886}{\scriptsize\,$\pm$0.015} & 0.902{\scriptsize\,$\pm$0.017} & 0.652{\scriptsize\,$\pm$0.012} & \underline{0.796}{\scriptsize\,$\pm$0.023} & \underline{0.822}{\scriptsize\,$\pm$0.021} & \underline{0.791}{\scriptsize\,$\pm$0.021} & 0.769{\scriptsize\,$\pm$0.004} \\
\cmidrule(lr){3-11}
 &  & Ministral-14B comp. & Full & 0.878{\scriptsize\,$\pm$0.015} & 0.893{\scriptsize\,$\pm$0.018} & 0.633{\scriptsize\,$\pm$0.013} & 0.770{\scriptsize\,$\pm$0.024} & 0.789{\scriptsize\,$\pm$0.022} & 0.777{\scriptsize\,$\pm$0.022} & 0.754{\scriptsize\,$\pm$0.004} \\
\midrule
\multirow{21}{*}{gpt-oss-120B} & \multirow{6}{*}{Qwen3.5-0.8B} & \multirow{2}{*}{Raw traces} & LoRA & \textbf{0.704}{\scriptsize\,$\pm$0.021} & \textbf{0.801}{\scriptsize\,$\pm$0.023} & \textbf{0.444}{\scriptsize\,$\pm$0.013} & \textbf{0.644}{\scriptsize\,$\pm$0.027} & \textbf{0.637}{\scriptsize\,$\pm$0.026} & \textbf{0.581}{\scriptsize\,$\pm$0.026} & \textbf{0.569}{\scriptsize\,$\pm$0.005} \\
 &  &  & Full & \textbf{0.734}{\scriptsize\,$\pm$0.021} & \textbf{0.765}{\scriptsize\,$\pm$0.024} & \textbf{0.481}{\scriptsize\,$\pm$0.013} & \textbf{0.681}{\scriptsize\,$\pm$0.026} & \textbf{0.633}{\scriptsize\,$\pm$0.026} & \textbf{0.639}{\scriptsize\,$\pm$0.025} & \textbf{0.577}{\scriptsize\,$\pm$0.005} \\
\cmidrule(lr){3-11}
 &  & \multirow{2}{*}{Llama-70B comp.} & LoRA & \underline{0.531}{\scriptsize\,$\pm$0.023} & \underline{0.718}{\scriptsize\,$\pm$0.026} & \underline{0.399}{\scriptsize\,$\pm$0.013} & \underline{0.568}{\scriptsize\,$\pm$0.028} & 0.547{\scriptsize\,$\pm$0.027} & \underline{0.544}{\scriptsize\,$\pm$0.026} & \underline{0.510}{\scriptsize\,$\pm$0.005} \\
 &  &  & Full & \underline{0.578}{\scriptsize\,$\pm$0.023} & \underline{0.726}{\scriptsize\,$\pm$0.026} & \underline{0.438}{\scriptsize\,$\pm$0.013} & \underline{0.634}{\scriptsize\,$\pm$0.027} & 0.534{\scriptsize\,$\pm$0.027} & \underline{0.567}{\scriptsize\,$\pm$0.026} & \underline{0.519}{\scriptsize\,$\pm$0.005} \\
\cmidrule(lr){3-11}
 &  & \multirow{2}{*}{Ministral-14B comp.} & LoRA & 0.476{\scriptsize\,$\pm$0.023} & 0.714{\scriptsize\,$\pm$0.026} & 0.397{\scriptsize\,$\pm$0.013} & 0.557{\scriptsize\,$\pm$0.028} & \underline{0.549}{\scriptsize\,$\pm$0.027} & 0.538{\scriptsize\,$\pm$0.026} & 0.507{\scriptsize\,$\pm$0.005} \\
 &  &  & Full & 0.537{\scriptsize\,$\pm$0.023} & 0.696{\scriptsize\,$\pm$0.026} & 0.417{\scriptsize\,$\pm$0.013} & 0.596{\scriptsize\,$\pm$0.027} & \underline{0.549}{\scriptsize\,$\pm$0.027} & 0.533{\scriptsize\,$\pm$0.026} & 0.516{\scriptsize\,$\pm$0.005} \\
\cmidrule(lr){2-11}
 & \multirow{6}{*}{Llama-3.1-8B} & \multirow{2}{*}{Raw traces} & LoRA & \textbf{0.690}{\scriptsize\,$\pm$0.022} & \textbf{0.839}{\scriptsize\,$\pm$0.021} & \textbf{0.639}{\scriptsize\,$\pm$0.013} & \textbf{0.746}{\scriptsize\,$\pm$0.024} & \textbf{0.631}{\scriptsize\,$\pm$0.026} & \textbf{0.754}{\scriptsize\,$\pm$0.022} & \textbf{0.704}{\scriptsize\,$\pm$0.004} \\
 &  &  & Full & \textbf{0.812}{\scriptsize\,$\pm$0.018} & \textbf{0.885}{\scriptsize\,$\pm$0.018} & \textbf{0.695}{\scriptsize\,$\pm$0.012} & \textbf{0.782}{\scriptsize\,$\pm$0.023} & \textbf{0.703}{\scriptsize\,$\pm$0.025} & \textbf{0.797}{\scriptsize\,$\pm$0.021} & \textbf{0.754}{\scriptsize\,$\pm$0.004} \\
\cmidrule(lr){3-11}
 &  & \multirow{2}{*}{Llama-70B comp.} & LoRA & \underline{0.632}{\scriptsize\,$\pm$0.023} & \underline{0.762}{\scriptsize\,$\pm$0.024} & \underline{0.580}{\scriptsize\,$\pm$0.013} & \underline{0.698}{\scriptsize\,$\pm$0.026} & \underline{0.575}{\scriptsize\,$\pm$0.027} & \underline{0.685}{\scriptsize\,$\pm$0.024} & \underline{0.631}{\scriptsize\,$\pm$0.005} \\
 &  &  & Full & \underline{0.709}{\scriptsize\,$\pm$0.021} & 0.809{\scriptsize\,$\pm$0.023} & \underline{0.626}{\scriptsize\,$\pm$0.013} & \underline{0.764}{\scriptsize\,$\pm$0.024} & \underline{0.597}{\scriptsize\,$\pm$0.027} & 0.721{\scriptsize\,$\pm$0.023} & 0.669{\scriptsize\,$\pm$0.005} \\
\cmidrule(lr){3-11}
 &  & \multirow{2}{*}{Ministral-14B comp.} & LoRA & 0.554{\scriptsize\,$\pm$0.023} & 0.755{\scriptsize\,$\pm$0.025} & 0.562{\scriptsize\,$\pm$0.013} & 0.677{\scriptsize\,$\pm$0.026} & 0.541{\scriptsize\,$\pm$0.027} & 0.677{\scriptsize\,$\pm$0.024} & 0.628{\scriptsize\,$\pm$0.005} \\
 &  &  & Full & 0.671{\scriptsize\,$\pm$0.022} & \underline{0.821}{\scriptsize\,$\pm$0.022} & 0.611{\scriptsize\,$\pm$0.013} & 0.749{\scriptsize\,$\pm$0.024} & 0.593{\scriptsize\,$\pm$0.027} & \underline{0.728}{\scriptsize\,$\pm$0.023} & \underline{0.671}{\scriptsize\,$\pm$0.005} \\
\cmidrule(lr){2-11}
 & \multirow{6}{*}{Qwen3.5-9B} & \multirow{2}{*}{Raw traces} & LoRA & \textbf{0.956}{\scriptsize\,$\pm$0.010} & \textbf{0.945}{\scriptsize\,$\pm$0.013} & \textbf{0.774}{\scriptsize\,$\pm$0.011} & \textbf{0.824}{\scriptsize\,$\pm$0.021} & \textbf{0.932}{\scriptsize\,$\pm$0.014} & \textbf{0.863}{\scriptsize\,$\pm$0.018} & \textbf{0.866}{\scriptsize\,$\pm$0.003} \\
 &  &  & Full & \textbf{0.961}{\scriptsize\,$\pm$0.009} & \textbf{0.948}{\scriptsize\,$\pm$0.013} & \textbf{0.771}{\scriptsize\,$\pm$0.011} & \textbf{0.830}{\scriptsize\,$\pm$0.021} & \textbf{0.932}{\scriptsize\,$\pm$0.014} & \textbf{0.865}{\scriptsize\,$\pm$0.018} & \textbf{0.866}{\scriptsize\,$\pm$0.003} \\
\cmidrule(lr){3-11}
 &  & \multirow{2}{*}{Llama-70B comp.} & LoRA & \underline{0.906}{\scriptsize\,$\pm$0.014} & 0.930{\scriptsize\,$\pm$0.015} & \underline{0.715}{\scriptsize\,$\pm$0.012} & \underline{0.790}{\scriptsize\,$\pm$0.023} & 0.865{\scriptsize\,$\pm$0.019} & \underline{0.843}{\scriptsize\,$\pm$0.019} & \underline{0.829}{\scriptsize\,$\pm$0.004} \\
 &  &  & Full & \underline{0.907}{\scriptsize\,$\pm$0.014} & \underline{0.932}{\scriptsize\,$\pm$0.014} & 0.718{\scriptsize\,$\pm$0.012} & 0.821{\scriptsize\,$\pm$0.022} & \underline{0.881}{\scriptsize\,$\pm$0.018} & \underline{0.831}{\scriptsize\,$\pm$0.019} & \underline{0.830}{\scriptsize\,$\pm$0.004} \\
\cmidrule(lr){3-11}
 &  & \multirow{2}{*}{Ministral-14B comp.} & LoRA & 0.897{\scriptsize\,$\pm$0.014} & \underline{0.934}{\scriptsize\,$\pm$0.014} & 0.702{\scriptsize\,$\pm$0.012} & 0.784{\scriptsize\,$\pm$0.023} & \underline{0.872}{\scriptsize\,$\pm$0.018} & 0.817{\scriptsize\,$\pm$0.020} & 0.816{\scriptsize\,$\pm$0.004} \\
 &  &  & Full & 0.896{\scriptsize\,$\pm$0.014} & 0.931{\scriptsize\,$\pm$0.015} & \underline{0.720}{\scriptsize\,$\pm$0.012} & \underline{0.823}{\scriptsize\,$\pm$0.021} & 0.865{\scriptsize\,$\pm$0.019} & 0.814{\scriptsize\,$\pm$0.020} & 0.817{\scriptsize\,$\pm$0.004} \\
\cmidrule(lr){2-11}
 & \multirow{3}{*}{gpt-oss-20B} & Raw traces & Full & \textbf{0.957}{\scriptsize\,$\pm$0.010} & \textbf{0.945}{\scriptsize\,$\pm$0.013} & \textbf{0.751}{\scriptsize\,$\pm$0.011} & \textbf{0.819}{\scriptsize\,$\pm$0.022} & \textbf{0.933}{\scriptsize\,$\pm$0.014} & \textbf{0.856}{\scriptsize\,$\pm$0.018} & \textbf{0.844}{\scriptsize\,$\pm$0.004} \\
\cmidrule(lr){3-11}
 &  & Llama-70B comp. & Full & \underline{0.904}{\scriptsize\,$\pm$0.014} & \underline{0.903}{\scriptsize\,$\pm$0.017} & \underline{0.680}{\scriptsize\,$\pm$0.012} & \underline{0.771}{\scriptsize\,$\pm$0.024} & 0.812{\scriptsize\,$\pm$0.021} & \underline{0.797}{\scriptsize\,$\pm$0.021} & \underline{0.776}{\scriptsize\,$\pm$0.004} \\
\cmidrule(lr){3-11}
 &  & Ministral-14B comp. & Full & 0.876{\scriptsize\,$\pm$0.016} & 0.894{\scriptsize\,$\pm$0.018} & 0.668{\scriptsize\,$\pm$0.012} & 0.759{\scriptsize\,$\pm$0.024} & \underline{0.813}{\scriptsize\,$\pm$0.021} & 0.783{\scriptsize\,$\pm$0.021} & 0.767{\scriptsize\,$\pm$0.004} \\
\bottomrule
\end{tabular}
\end{adjustbox}

  \caption{Downstream evaluation accuracy across the 48 main-grid runs plus seven Qwen-teacher truncation ablations. ``Overall'' is averaged across every evaluated record; bold marks the within-(student, method) winner per column and underline marks the second best result. Entries excluded by design (gpt-oss-20B LoRA under either teacher; answer-only for gpt-oss-20B or under the gpt-oss teacher; truncation under the gpt-oss teacher) render as ``---''.}
  \label{tab:eval_results_full}
\end{table*}

\FloatBarrier

% ===================================================================
\section{Length-Matched Truncation Ablation}
\label{app:truncated_ablation}
% ===================================================================

Table~\ref{tab:truncated_ablation} reports the Qwen-teacher truncation ablation used in Section~\ref{sec:results_eval}. Trunc. cuts the raw trace to the same per-example token length as the Ministral-14B-compressed trace; L70 and M14 are model-compressed traces.

\begin{table}[h]
  \centering
  \footnotesize
  \begin{tabular}{llccc}
    \toprule
    \textbf{Student} & \textbf{Method} & \textbf{Trunc.} & \textbf{L70} & \textbf{M14} \\
    \midrule
    Qwen-0.8B & LoRA & 0.479 & \textbf{0.506} & 0.500 \\
    Qwen-0.8B & Full & 0.487 & 0.481 & \textbf{0.500} \\
    Llama-8B & LoRA & 0.417 & 0.620 & \textbf{0.621} \\
    Llama-8B & Full & 0.596 & \textbf{0.665} & 0.652 \\
    Qwen-9B & LoRA & 0.826 & \textbf{0.830} & 0.815 \\
    Qwen-9B & Full & 0.830 & \textbf{0.834} & 0.817 \\
    gpt-oss-20B & Full & \textbf{0.769} & \textbf{0.769} & 0.754 \\
    \bottomrule
  \end{tabular}
  \caption{Qwen-teacher length-reduced ablation. Bold marks the best non-raw source in each row.}
  \label{tab:truncated_ablation}
\end{table}

\FloatBarrier

% ===================================================================
\section{Inference-Time Efficiency}
\label{app:eval_efficiency}
% ===================================================================

Table~\ref{tab:eval_efficiency} reports inference-time efficiency statistics for the 48 main-grid runs plus seven Qwen-teacher truncation ablations.

\begin{table*}[h]
  \centering
  \begin{adjustbox}{max width=\textwidth}
\begin{tabular}{llllrrrrr}
\toprule
\multirow{2}{*}{\textbf{Generator}} & \multirow{2}{*}{\textbf{Student}} & \multirow{2}{*}{\textbf{Data}} & \multirow{2}{*}{\textbf{Method}} & \textbf{Avg.\ Compl.} & \textbf{Avg.\ Compl.} & \textbf{Trunc.} & \textbf{Med.\ Reason} & \textbf{Acc / 1k} \\
& & & & \textbf{Tokens (all)} & \textbf{Tokens (excl.\ trunc.)} & \textbf{Rate} & \textbf{Chars} & \textbf{Tokens} \\
\midrule
\multirow{34}{*}{Qwen3.5-397B} & \multirow{10}{*}{Qwen3.5-0.8B} & \multirow{2}{*}{Raw traces} & LoRA & 2,954 & 2,461 & 0.086 & 8,999 & 0.1786 \\
 &  &  & Full & 2,813 & 2,406 & 0.070 & 8,791 & 0.1893 \\
\cmidrule(lr){3-9}
 &  & \multirow{2}{*}{Answer-Only} & LoRA & 4 & 4 & 0.000 & 0 & 93.1334 \\
 &  &  & Full & 8,189 & 4,840 & 0.999 & 0 & 0.0001 \\
\cmidrule(lr){3-9}
 &  & \multirow{2}{*}{Truncated} & LoRA & 321 & 321 & 0.000 & 1,181 & 1.4943 \\
 &  &  & Full & 278 & 278 & 0.000 & 1,031 & 1.7537 \\
\cmidrule(lr){3-9}
 &  & \multirow{2}{*}{Llama-70B comp.} & LoRA & 234 & 233 & 0.000 & 1,054 & 2.1647 \\
 &  &  & Full & 2,097 & 267 & 0.231 & 1,080 & 0.2293 \\
\cmidrule(lr){3-9}
 &  & \multirow{2}{*}{Ministral-14B comp.} & LoRA & 179 & 178 & 0.000 & 736 & 2.7865 \\
 &  &  & Full & 178 & 178 & 0.000 & 733 & 2.8027 \\
\cmidrule(lr){2-9}
 & \multirow{10}{*}{Llama-3.1-8B} & \multirow{2}{*}{Raw traces} & LoRA & 3,528 & 2,676 & 0.154 & 12,973 & 0.1861 \\
 &  &  & Full & 2,622 & 2,263 & 0.061 & 7,783 & 0.2727 \\
\cmidrule(lr){3-9}
 &  & \multirow{2}{*}{Answer-Only} & LoRA & 2 & 2 & 0.000 & 0 & 293.3698 \\
 &  &  & Full & 2 & 2 & 0.000 & 0 & 90.4606 \\
\cmidrule(lr){3-9}
 &  & \multirow{2}{*}{Truncated} & LoRA & 1,328 & 773 & 0.075 & 2,773 & 0.3142 \\
 &  &  & Full & 184 & 158 & 0.003 & 1 & 3.2319 \\
\cmidrule(lr){3-9}
 &  & \multirow{2}{*}{Llama-70B comp.} & LoRA & 233 & 232 & 0.000 & 1,085 & 2.6613 \\
 &  &  & Full & 232 & 226 & 0.001 & 1,051 & 2.8686 \\
\cmidrule(lr){3-9}
 &  & \multirow{2}{*}{Ministral-14B comp.} & LoRA & 180 & 175 & 0.001 & 745 & 3.4487 \\
 &  &  & Full & 166 & 165 & 0.000 & 714 & 3.9282 \\
\cmidrule(lr){2-9}
 & \multirow{10}{*}{Qwen3.5-9B} & \multirow{2}{*}{Raw traces} & LoRA & 1,827 & 1,714 & 0.018 & 4,219 & 0.4717 \\
 &  &  & Full & 1,814 & 1,757 & 0.009 & 4,180 & 0.4775 \\
\cmidrule(lr){3-9}
 &  & \multirow{2}{*}{Answer-Only} & LoRA & 4 & 4 & 0.000 & 0 & 192.7208 \\
 &  &  & Full & 8,188 & 3,900 & 0.999 & 0 & 0.0672 \\
\cmidrule(lr){3-9}
 &  & \multirow{2}{*}{Truncated} & LoRA & 483 & 482 & 0.000 & 1,668 & 1.7106 \\
 &  &  & Full & 370 & 369 & 0.000 & 1,318 & 2.2436 \\
\cmidrule(lr){3-9}
 &  & \multirow{2}{*}{Llama-70B comp.} & LoRA & 223 & 222 & 0.000 & 1,004 & 3.7142 \\
 &  &  & Full & 223 & 223 & 0.000 & 1,004 & 3.7377 \\
\cmidrule(lr){3-9}
 &  & \multirow{2}{*}{Ministral-14B comp.} & LoRA & 168 & 167 & 0.000 & 700 & 4.8637 \\
 &  &  & Full & 232 & 185 & 0.006 & 685 & 3.5236 \\
\cmidrule(lr){2-9}
 & \multirow{4}{*}{gpt-oss-20B} & Raw traces & Full & 2,261 & 2,068 & 0.032 & 6,539 & 0.3606 \\
\cmidrule(lr){3-9}
 &  & Truncated & Full & 319 & 319 & 0.000 & 1,089 & 2.4101 \\
\cmidrule(lr){3-9}
 &  & Llama-70B comp. & Full & 323 & 228 & 0.012 & 1,046 & 2.3798 \\
\cmidrule(lr){3-9}
 &  & Ministral-14B comp. & Full & 468 & 176 & 0.036 & 742 & 1.6104 \\
\midrule
\multirow{21}{*}{gpt-oss-120B} & \multirow{6}{*}{Qwen3.5-0.8B} & \multirow{2}{*}{Raw traces} & LoRA & 1,394 & 1,200 & 0.028 & 3,478 & 0.4083 \\
 &  &  & Full & 1,428 & 1,256 & 0.025 & 3,696 & 0.4043 \\
\cmidrule(lr){3-9}
 &  & \multirow{2}{*}{Llama-70B comp.} & LoRA & 220 & 219 & 0.000 & 1,020 & 2.3182 \\
 &  &  & Full & 292 & 217 & 0.009 & 997 & 1.7781 \\
\cmidrule(lr){3-9}
 &  & \multirow{2}{*}{Ministral-14B comp.} & LoRA & 166 & 166 & 0.000 & 663 & 3.0491 \\
 &  &  & Full & 170 & 170 & 0.000 & 687 & 3.0321 \\
\cmidrule(lr){2-9}
 & \multirow{6}{*}{Llama-3.1-8B} & \multirow{2}{*}{Raw traces} & LoRA & 1,586 & 1,284 & 0.044 & 3,572 & 0.4442 \\
 &  &  & Full & 1,237 & 1,106 & 0.018 & 3,263 & 0.6096 \\
\cmidrule(lr){3-9}
 &  & \multirow{2}{*}{Llama-70B comp.} & LoRA & 199 & 199 & 0.000 & 934 & 3.1629 \\
 &  &  & Full & 193 & 192 & 0.000 & 914 & 3.4656 \\
\cmidrule(lr){3-9}
 &  & \multirow{2}{*}{Ministral-14B comp.} & LoRA & 169 & 168 & 0.000 & 717 & 3.7270 \\
 &  &  & Full & 151 & 150 & 0.000 & 640 & 4.4296 \\
\cmidrule(lr){2-9}
 & \multirow{6}{*}{Qwen3.5-9B} & \multirow{2}{*}{Raw traces} & LoRA & 959 & 900 & 0.008 & 2,488 & 0.9032 \\
 &  &  & Full & 924 & 879 & 0.006 & 2,459 & 0.9372 \\
\cmidrule(lr){3-9}
 &  & \multirow{2}{*}{Llama-70B comp.} & LoRA & 194 & 194 & 0.000 & 866 & 4.2715 \\
 &  &  & Full & 192 & 191 & 0.000 & 831 & 4.3169 \\
\cmidrule(lr){3-9}
 &  & \multirow{2}{*}{Ministral-14B comp.} & LoRA & 153 & 153 & 0.000 & 623 & 5.3482 \\
 &  &  & Full & 183 & 158 & 0.003 & 612 & 4.4642 \\
\cmidrule(lr){2-9}
 & \multirow{3}{*}{gpt-oss-20B} & Raw traces & Full & 958 & 910 & 0.007 & 2,649 & 0.8802 \\
\cmidrule(lr){3-9}
 &  & Llama-70B comp. & Full & 253 & 198 & 0.007 & 877 & 3.0720 \\
\cmidrule(lr){3-9}
 &  & Ministral-14B comp. & Full & 416 & 161 & 0.032 & 672 & 1.8454 \\
\bottomrule
\end{tabular}
\end{adjustbox}

  \caption{Inference-time efficiency. Truncation rate is the fraction of records hitting the 8{,}192-token cap; ``Acc / 1k Tokens'' is overall accuracy divided by mean completion tokens (in thousands).}
  \label{tab:eval_efficiency}
\end{table*}

\FloatBarrier

% ===================================================================
\section{Faithfulness of Compressed Traces}
\label{app:faithfulness}
% ===================================================================

Section~\ref{sec:method} treats the compressor $\mathcal{M}_C$ as a black box: it takes $(q_i, t_i, a_i)$ and returns a shorter $\hat{t}_i$ that is assumed to preserve the answer-bearing reasoning. Because the canonical answer $a_i$ is verified against $a_i^*$ before compression and re-appended unchanged after, a compressor that silently corrupts the reasoning---merging two derivation steps that should have remained separate, hallucinating an intermediate constant, dropping a case in a case split---would still produce a syntactically well-formed training example, and the student would learn to imitate the corruption. The downstream-accuracy gap between raw and compressed runs is an indirect upper bound on the cost of such corruptions (Section~\ref{sec:limitations}), but it conflates ``the trace lost a load-bearing step'' with ``the student failed to imitate the compressed format''. This appendix describes the separate, trace-level check we run to disentangle the two and reports what it finds.

\subsection{Method}
\label{app:faithfulness_method}

For each compressed trace $\hat{t}_i$ produced in Stage~2, we re-issue the triple $(q_i, t_i, \hat{t}_i, a_i^*)$ to a third LLM, the \emph{judge}. The judge is asked to compare $\hat{t}_i$ against $t_i$ along three axes and to emit a structured verdict:
\begin{itemize}[leftmargin=*,topsep=2pt,itemsep=1pt]
  \item \textbf{Faithfulness} (1--5): is every claim, calculation, and conclusion in $\hat{t}_i$ supported by or consistent with $t_i$? A faithful compression may omit redundant exploration, false starts, and verbosity, but it must not introduce unsupported claims.
  \item \textbf{Coverage} (1--5): are the essential reasoning steps needed to support $a_i^*$ preserved in $\hat{t}_i$?
  \item \textbf{Clarity} (1--5): is the compressed trace coherent and usable as a deliberation?
  \item \textbf{Verdict}: one of \texttt{faithful}, \texttt{partially\_faithful}, \texttt{unfaithful}.
  \item \textbf{Answer alignment}: a Boolean indicating whether $\hat{t}_i$, read on its own, would support the canonical answer $a_i^*$.
  \item \textbf{Failure modes}: a (possibly empty) subset of \{contradiction, unsupported\_claim, omitted\_key\_step, answer\_mismatch, overcompression, incoherent, empty\_original, empty\_compressed\}.
  \item \textbf{Rationale}: a short free-text justification.
\end{itemize}
The verdict is constrained server-side by a JSON schema (\texttt{response\_format} with a Pydantic-derived spec), so structural validation is essentially free; the only realistic parse failure is a response truncated at the judge's \texttt{max\_completion\_tokens}. The full prompt includes the question, the canonical answer, both traces (middle-truncated to 40k and 12k characters respectively if longer, so the judge always sees both ends), and an explicit instruction that omissions of redundant exploration are acceptable but unsupported claims are not. We use \texttt{gpt-oss-120B} as the judge using structured decoding to ensure predictable output that can be easily parsed.

\subsection{Results}
\label{app:faithfulness_results}

Table~\ref{tab:faithfulness} reports the judge's verdict distribution per (compressor, domain, dataset). For each compressor we report the number of traces assessed ($N$), mean faithfulness score (1--5), the percentage rated \texttt{faithful} (verdict, not score), and the percentage with positive \texttt{answer\_alignment}.

\begin{table*}[t]
  \centering
  \begin{adjustbox}{max width=\textwidth}
\begin{tabular}{ll rrrr rrrr}
\toprule
 &  & \multicolumn{4}{c}{\textbf{Llama-3.3-70B}} & \multicolumn{4}{c}{\textbf{Ministral-3-14B}} \\
\cmidrule(lr){3-6} \cmidrule(lr){7-10}
\textbf{Domain} & \textbf{Dataset} & \textbf{N} & \textbf{Faith.} & \textbf{\%~Faithful} & \textbf{\%~Align} & \textbf{N} & \textbf{Faith.} & \textbf{\%~Faithful} & \textbf{\%~Align} \\
\midrule
\multirow{4}{*}{Math} & AQUA-RAT & 87,441 & 4.97 & 98.3 & 97.3 & 87,441 & 4.95 & 97.8 & 96.5 \\
 & GSM8k & 7,186 & 5.00 & 99.7 & 85.8 & 7,186 & 4.99 & 99.6 & 85.4 \\
 & MultiArith & 416 & 4.97 & 98.8 & 88.7 & 416 & 4.99 & 99.8 & 90.9 \\
 & SVAMP & 960 & 4.99 & 99.6 & 92.3 & 960 & 4.99 & 99.4 & 94.2 \\
\midrule
\multirow{2}{*}{Science} & ARC-Challenge & 1,079 & 4.99 & 95.8 & 97.6 & 1,079 & 4.97 & 97.0 & 97.0 \\
 & GPQA Diamond & 135 & 4.76 & 90.4 & 92.6 & 135 & 4.75 & 92.6 & 94.8 \\
\midrule
\multirow{1}{*}{Logic} & CommonsenseQA & 8,593 & 4.98 & 96.0 & 98.2 & 8,593 & 4.97 & 97.9 & 98.4 \\
\midrule
\multirow{3}{*}{Medical} & MedExpQA & 395 & 4.98 & 98.2 & 99.0 & 395 & 4.96 & 98.2 & 98.7 \\
 & MedMCQA & 167,277 & 4.98 & 97.9 & 97.9 & 167,277 & 4.97 & 98.0 & 97.9 \\
 & MedQA & 9,778 & 4.98 & 98.7 & 99.1 & 9,778 & 4.97 & 98.3 & 99.3 \\
\midrule
\multicolumn{2}{l}{\textbf{Overall}} & \textbf{283,260} & \textbf{4.98} & \textbf{98.1} & \textbf{97.4} & \textbf{283,260} & \textbf{4.97} & \textbf{98.0} & \textbf{97.2} \\
\bottomrule
\end{tabular}
\end{adjustbox}

  \caption{LLM-as-judge faithfulness assessment of compressed traces, scored by \texttt{gpt-oss-120B}. \emph{Faith.} is the mean faithfulness score (1--5). \emph{\%~Faithful} is the percentage assigned verdict \texttt{faithful} (the remainder is split between \texttt{partially\_faithful} and \texttt{unfaithful}). \emph{\%~Align} is the percentage with positive answer alignment: the judge, given $a_i^*$, agrees that the compressed trace supports it.}
  \label{tab:faithfulness}
\end{table*}

\paragraph{Headline numbers.}
The judge rates the overwhelming majority of compressed traces as faithful (verdict-level, not score). Across the full $283{,}260$-trace corpus, the Llama-70B compressor scores $98.1\%$ \texttt{faithful} with mean faithfulness $4.98/5$ and $97.4\%$ positive answer alignment; the Ministral-14B compressor scores $98.0\%$ \texttt{faithful}, $4.97/5$, and $97.2\%$ alignment. Verdict-level rates are above $95\%$ in every (compressor, dataset) configuration except GPQA Diamond (90--93\%), where the small sample and the difficulty of the source material together explain the dip.

\paragraph{Failure-mode breakdown.}
The remaining $\sim$2\% of non-\texttt{faithful} verdicts are dominated by \texttt{omitted\_key\_step}, \texttt{answer\_mismatch}, and \texttt{unsupported\_claim}; \texttt{overcompression}, \texttt{contradiction}, and \texttt{incoherent} are rare. \texttt{empty\_original} / \texttt{empty\_compressed} are essentially zero, confirming that Stage~2 rarely degenerates to a degenerate output.

\paragraph{Interpretation.}
The $\sim$2\% non-\texttt{faithful} rate is small relative to the raw-vs-compressed accuracy gap reported in Section~\ref{sec:results_eval} ($\leq 8.5$ points at $\geq$8B), so the judge data is consistent with the limitations-section claim that propagated trace corruption is a small but non-zero contributor and not the dominant explanation for the gap. The remaining gap is then accounted for by the legitimate cost of stripping redundant deliberation that the student would otherwise have imitated.

% ===================================================================
\section{Training Stack and Compute}
\label{app:compute}
% ===================================================================

This appendix records the implementation and compute used to produce the 48 main-grid student runs, seven Qwen-teacher truncation ablations, and the upstream trace-generation and compression stages.

\subsection{Hardware}
\label{app:hardware}

All runs use NVIDIA H100 GPUs (SXM, 80\,GB) connected by NVLink/NVSwitch within a node and InfiniBand between nodes. Each student training run uses a single node; the per-job GPU count depends on the student and method (4 GPUs for the 0.8B student and for the 8B/9B LoRA-on-compressed runs that fit on 4 GPUs, 8 GPUs for the 8B/9B raw and full-FSDP runs and for every gpt-oss-20B run). The exact GPU count per run is recorded in Table~\ref{tab:compute_per_run}. Trace generation and trace compression use 8 H100s on a single node.

\subsection{Trace Generation and Compression Stack}
\label{app:vllm_stack}

Both upstream stages use vLLM with continuous batching and FP8 weights for the teacher.

\paragraph{Teacher (Stage~1).}
Qwen3.5-397B-A17B-FP8 served on 8$\times$H100 with tensor parallelism 8, data parallelism 1, max model length 16{,}384, GPU memory utilisation 0.92, and the \texttt{qwen3} reasoning parser. Sampling batch size is 512; we use rejection sampling with a single attempt per remaining question per round.

\paragraph{Compressor: Llama-3.3-70B-Instruct.}
Served on 8$\times$H100 with tensor parallelism 4 and data parallelism 2 (two replicas, each 4 GPUs).
Max model length 16{,}384, GPU memory utilisation 0.92, temperature 0.3, batch size 1{,}024.

\paragraph{Compressor: Ministral-3-14B-Instruct-2512.}
Served on 8$\times$H100 with tensor parallelism 1 and data parallelism 8 (one replica per GPU). Max model length 16{,}384, GPU memory utilisation 0.92, temperature 0.3, batch size 4{,}096.

\subsection{Student Training Stack}
\label{app:train_stack}

All student runs use Axolotl with the following common configuration: chat template inherited from the tokenizer, EOT token \texttt{<|im\_end|>}, \texttt{train\_on\_eot: turn}, \texttt{split\_thinking: true}, BF16 mixed precision with TF32 enabled, sequence length 16{,}384, sample packing with padding to sequence length, micro batch size 1, gradient accumulation 8, cosine schedule with 10\% warmup, FlashAttention~2, the CutCrossEntropy plugin, and the Liger plugin (fused RMSNorm, RoPE, GLU activation, and gated RMSNorm).

\paragraph{LoRA.}
Rank 64, $\alpha{=}32$, dropout 0.05, applied to \{q,k,v,o,gate,up,down\}\_proj.
Learning rate $1\!\times\!10^{-4}$, one epoch, optimizer \texttt{adamw\_bnb\_8bit}, gradient checkpointing on.

\paragraph{Full fine-tuning (0.8B).}
Learning rate $2\!\times\!10^{-5}$, one epoch, optimizer \texttt{adamw\_torch\_8bit}.

\paragraph{Full fine-tuning (8B / 9B / 20B).}
FSDP~v2 with full sharding (\texttt{FULL\_SHARD}), \texttt{TRANSFORMER\_BASED\_WRAP} auto-wrap (\texttt{Qwen3\_5DecoderLayer}, \texttt{LlamaDecoderLayer}, or \texttt{GptOssDecoderLayer}), reshard after forward, activation checkpointing, full state dict. Same learning rate and epoch count as the 0.8B full setting.

\paragraph{Answer-only ablation.}
The answer-only data source is produced by stripping the entire \think{}\ldots\thinkend{} block from each correct trace, leaving only the user question and the final answer. Training uses the same chat template, optimiser, schedule, and epoch count as the matching reasoning-trace configuration; the only difference is the training data.

\subsection{Compute per Run}
\label{app:compute_per_run}

Table~\ref{tab:compute_per_run} reports wall-clock time and GPU-hours per student run. Wall-clock numbers are taken from Table~\ref{tab:efficiency}; GPU-hours are wall-clock $\times$ per-job GPU count (4 or 8, depending on the run). Token counts are the source-dataset trainable-token totals (raw, compressed, truncated, or answer-only, before the epoch multiplier), matching the right half of Table~\ref{tab:efficiency}.

\begin{table*}[t]
  \centering
  \footnotesize
  \setlength{\tabcolsep}{4pt}
  \resizebox{\textwidth}{!}{%
    \begin{tabular}{lll l rrrr}
\toprule
\textbf{Generator} & \textbf{Student} & \textbf{Method} & \textbf{Data} & \textbf{GPUs} & \textbf{Time (h)} & \textbf{GPU-h} & \textbf{Trainable tokens (M)} \\
\midrule
  \multirow{34}{*}{Qwen3.5-397B} & \multirow{10}{*}{Qwen3.5-0.8B} & LoRA & Raw                    & 4 &  1.1 &   4.5 &   635 \\
   &  & LoRA & Answer-Only            & 4 &  0.1 &   0.3 &    25 \\
   &  & LoRA & Truncated              & 4 &  0.1 &   0.6 &    75 \\
   &  & LoRA & Llama-70B-compressed   & 4 &  0.2 &   0.8 &   102 \\
   &  & LoRA & Ministral-14B-comp.    & 4 &  0.1 &   0.6 &    74 \\
\cmidrule(lr){3-8}
   &  & Full & Raw                    & 4 &  1.1 &   4.2 &   635 \\
   &  & Full & Answer-Only            & 4 &  0.1 &   0.3 &    25 \\
   &  & Full & Truncated              & 4 &  0.1 &   0.6 &    75 \\
   &  & Full & Llama-70B-compressed   & 4 &  0.2 &   0.7 &   102 \\
   &  & Full & Ministral-14B-comp.    & 4 &  0.1 &   0.6 &    74 \\
\cmidrule(lr){2-8}
   & \multirow{10}{*}{Llama-3.1-8B} & LoRA & Raw                    & 8 &  2.7 &  21.8 &   589 \\
   &  & LoRA & Answer-Only            & 8 &  0.1 &   0.8 &    23 \\
   &  & LoRA & Truncated              & 8 &  0.3 &   2.3 &    71 \\
   &  & LoRA & Llama-70B-compressed   & 4 &  0.8 &   3.1 &    97 \\
   &  & LoRA & Ministral-14B-comp.    & 4 &  0.6 &   2.3 &    71 \\
\cmidrule(lr){3-8}
   &  & Full & Raw                    & 8 &  3.1 &  24.4 &   589 \\
   &  & Full & Answer-Only            & 8 &  0.1 &   1.0 &    23 \\
   &  & Full & Truncated              & 8 &  0.3 &   2.7 &    71 \\
   &  & Full & Llama-70B-compressed   & 4 &  0.9 &   3.4 &    97 \\
   &  & Full & Ministral-14B-comp.    & 4 &  0.6 &   2.5 &    71 \\
\cmidrule(lr){2-8}
   & \multirow{10}{*}{Qwen3.5-9B} & LoRA & Raw                    & 8 &  2.7 &  21.6 &   635 \\
   &  & LoRA & Answer-Only            & 8 &  0.1 &   1.0 &    24 \\
   &  & LoRA & Truncated              & 8 &  0.3 &   2.6 &    74 \\
   &  & LoRA & Llama-70B-compressed   & 4 &  0.8 &   3.3 &   102 \\
   &  & LoRA & Ministral-14B-comp.    & 4 &  0.6 &   2.5 &    74 \\
\cmidrule(lr){3-8}
   &  & Full & Raw                    & 8 &  3.3 &  26.7 &   635 \\
   &  & Full & Answer-Only            & 8 &  0.2 &   1.4 &    24 \\
   &  & Full & Truncated              & 8 &  0.4 &   3.3 &    74 \\
   &  & Full & Llama-70B-compressed   & 4 &  1.0 &   4.1 &   102 \\
   &  & Full & Ministral-14B-comp.    & 4 &  0.8 &   3.0 &    74 \\
\cmidrule(lr){2-8}
   & \multirow{4}{*}{gpt-oss-20B} & Full & Raw                    & 8 &  3.1 &  25.0 &   600 \\
   &  & Full & Truncated              & 8 &  0.5 &   3.9 &    88 \\
   &  & Full & Llama-70B-compressed   & 8 &  0.6 &   4.9 &   113 \\
   &  & Full & Ministral-14B-comp.    & 8 &  0.5 &   4.0 &    87 \\
\midrule
  \multirow{21}{*}{gpt-oss-120B} & \multirow{6}{*}{Qwen3.5-0.8B} & LoRA & Raw                    & 4 &  0.6 &   2.3 &   315 \\
   &  & LoRA & Llama-70B-compressed   & 4 &  0.2 &   0.6 &    81 \\
   &  & LoRA & Ministral-14B-comp.    & 4 &  0.1 &   0.6 &    69 \\
\cmidrule(lr){3-8}
   &  & Full & Raw                    & 4 &  0.5 &   2.2 &   315 \\
   &  & Full & Llama-70B-compressed   & 4 &  0.2 &   0.6 &    81 \\
   &  & Full & Ministral-14B-comp.    & 4 &  0.1 &   0.6 &    69 \\
\cmidrule(lr){2-8}
   & \multirow{6}{*}{Llama-3.1-8B} & LoRA & Raw                    & 8 &  1.4 &  11.0 &   298 \\
   &  & LoRA & Llama-70B-compressed   & 4 &  0.6 &   2.5 &    78 \\
   &  & LoRA & Ministral-14B-comp.    & 4 &  0.5 &   2.1 &    66 \\
\cmidrule(lr){3-8}
   &  & Full & Raw                    & 8 &  1.5 &  12.4 &   298 \\
   &  & Full & Llama-70B-compressed   & 4 &  0.7 &   2.7 &    78 \\
   &  & Full & Ministral-14B-comp.    & 4 &  0.6 &   2.3 &    66 \\
\cmidrule(lr){2-8}
   & \multirow{6}{*}{Qwen3.5-9B} & LoRA & Raw                    & 8 &  1.3 &  10.8 &   315 \\
   &  & LoRA & Llama-70B-compressed   & 4 &  0.7 &   2.7 &    81 \\
   &  & LoRA & Ministral-14B-comp.    & 4 &  0.6 &   2.3 &    69 \\
\cmidrule(lr){3-8}
   &  & Full & Raw                    & 8 &  1.7 &  13.3 &   315 \\
   &  & Full & Llama-70B-compressed   & 4 &  0.8 &   3.3 &    81 \\
   &  & Full & Ministral-14B-comp.    & 4 &  0.7 &   2.9 &    69 \\
\cmidrule(lr){2-8}
   & \multirow{3}{*}{gpt-oss-20B} & Full & Raw                    & 8 &  1.6 &  13.1 &   310 \\
   &  & Full & Llama-70B-compressed   & 8 &  0.5 &   4.1 &    94 \\
   &  & Full & Ministral-14B-comp.    & 8 &  0.5 &   3.7 &    83 \\
\bottomrule
\end{tabular}
  }
  \caption{Per-run wall-clock time, GPU count, GPU-hours, and source-dataset trainable tokens for the 48 main-grid student runs plus seven Qwen-teacher truncation ablations. Wall-clock and token totals are reproduced from Table~\ref{tab:efficiency}; GPU-hours are wall-clock $\times$ GPU count.}
  \label{tab:compute_per_run}
\end{table*}

\paragraph{Aggregate compute.}
Summed across the 55 student runs, training consumes $\approx$\,280.9 H100-hours of pure GPU time. The seven truncation ablations add 16.0 H100-h over the 264.9 H100-h main grid. Within each student, the raw-trace and full-FSDP runs dominate (e.g., Qwen3.5-9B raw under the Qwen teacher alone accounts for 26.7 H100-h), and the answer-only, truncated, and compressed runs together account for less than a third of the per-student total. Trace generation and compression still dominate the project budget: the teacher pass over 345{,}383 questions and the two compressor passes over the 283{,}335 / 281{,}911 correct traces from each teacher each occupy an 8$\times$H100 node.

\paragraph{Upstream wall-clock.}
On the same 8$\times$H100 node, the teacher pass over 345{,}383 questions took $\approx$\,43.5\,h ($\approx$\,348\,H100-h); compressing the 283{,}335 correct traces took 1\,h\,09\,m with Ministral-3-14B ($\approx$\,9.2\,H100-h) and 8\,h\,19\,m with Llama-3.3-70B ($\approx$\,66.5\,H100-h), for $\approx$\,75.7\,H100-h of compression in total. The upstream stages thus contribute $\approx$\,424\,H100-hours, roughly $1.5\times$ the student-training total of 280.9\,H100-hours.

\subsection{Energy and Carbon Footprint}
\label{app:carbon}

Student training was performed on H100 GPUs hosted in a Finnish datacentre.
We estimate energy and carbon at the GPU-hour level using the per-GPU SXM TDP (700~W) multiplied by an assumed datacentre PUE of 1.15, giving an effective draw of 805~W per H100-hour including cooling and infrastructure overhead.

\paragraph{Estimated emissions.}
Student training plus upstream trace generation and compression amount to $704.9 \times 0.805 \approx 567.4$~kWh. We use 79~gCO\textsubscript{2}eq/kWh, the published Finnish annual average for 2023 (Fingrid / IEA) where the servers are hosted; 2024 was slightly lower. We estimate a total of $567.4 \times 0.079 \approx 44.8$~kgCO\textsubscript{2}eq.

\end{document}